\documentclass{article}
\PassOptionsToPackage{numbers, compress}{natbib}

\usepackage[preprint]{neurips_data_2024}
\usepackage{times}
\usepackage{latexsym}
\usepackage[T1]{fontenc}
\usepackage[utf8]{inputenc}
\usepackage{microtype}
\usepackage{inconsolata}
\usepackage{algorithm}
\usepackage{algorithmic}
\usepackage{CJKutf8}

\usepackage{color}

\usepackage{hyperref}       
\usepackage{url}            
\usepackage{booktabs}       
\usepackage{amsfonts}       
\usepackage{nicefrac}       
\usepackage{microtype}      
\usepackage{xcolor}         
\usepackage{stfloats}
\usepackage{enumitem}
\usepackage{multirow}
\usepackage{graphicx}
\usepackage{setspace}
\usepackage{algorithm}
\usepackage{algorithmic}
\usepackage{bbm}
\usepackage{caption}
\usepackage{subfigure}
\usepackage{tcolorbox}
\usepackage{textcomp} 
\usepackage{array}
\usepackage{caption}
\usepackage[utf8]{inputenc} 
\usepackage[T1]{fontenc}    
\usepackage{hyperref}       
\usepackage{url}            
\usepackage{booktabs}       
\usepackage{amsfonts}       
\usepackage{nicefrac}       
\usepackage{microtype}      
\usepackage{xcolor}         
\usepackage{wrapfig} 
\usepackage{subcaption}

\title{AI Hospital: Benchmarking Large Language Models in a Multi-agent Medical Interaction Simulator}

\author{
  Zhihao Fan$^{1}$,
  Jialong Tang$^{1}$,
  Wei Chen$^{2}$\thanks{Corresponding author. Email: lemuria\_chen@hust.edu.cn}, \hspace{0.1em}   
  Siyuan Wang$^{3}$,
  Zhongyu Wei$^{3}$,
  \vspace{0.1em}   
  \\ \textbf{
  Jun Xie$^{1}$,
  Fei Huang$^{1}$,
  Jingren Zhou$^{1}$}
  \\   \vspace{0.1em}   
  $^1$Alibaba Inc., $^2$Huazhong University of Science and Technology, $^3$Fudan University \\
 \vspace{0.1em}   
  $^1$fanzhihao.fzh@alibaba-inc.com
}

\begin{document}
\maketitle

\maketitle

\begin{abstract}
Artificial intelligence has significantly advanced healthcare, particularly through large language models (LLMs) that excel in medical question answering benchmarks. However, their real-world clinical application remains limited due to the complexities of doctor-patient interactions. To address this, we introduce \textbf{AI Hospital}, a multi-agent framework simulating dynamic medical interactions between \emph{Doctor} as player and NPCs including \emph{Patient}, \emph{Examiner}, \emph{Chief Physician}. This setup allows for realistic assessments of LLMs in clinical scenarios. We develop the Multi-View Medical Evaluation (MVME) benchmark, utilizing high-quality Chinese medical records and NPCs to evaluate LLMs' performance in symptom collection, examination recommendations, and diagnoses. Additionally, a dispute resolution collaborative mechanism is proposed to enhance diagnostic accuracy through iterative discussions. Despite improvements, current LLMs exhibit significant performance gaps in multi-turn interactions compared to one-step approaches. Our findings highlight the need for further research to bridge these gaps and improve LLMs' clinical diagnostic capabilities. Our data, code, and experimental results are all open-sourced at \url{https://github.com/LibertFan/AI_Hospital}. 
\end{abstract}


\section{Introduction}
Healthcare has seen significant progress with artificial intelligence in recent years~\cite{bajwa2021artificial}, especially through the development of large language models (LLMs)~\cite{OpenAI2023GPT4TR,touvron2023llama,bai2023qwen}. These models demonstrate impressive performance on static medical question answering benchmarks like MedQA~\cite{jin2021disease}, PubMedQA~\cite{jin2019pubmedqa}, and MedMCQA~\cite{pal2022medmcqa}, even rivaling human experts. However, a significant gap remains between LLMs performance on these benchmarks and their real-world application in clinical diagnosis.  

In practice, patients frequently lack sufficient medical knowledge and may possess ambiguous understandings of various medical concepts, making it difficult for them to accurately and comprehensively communicate their physical condition to healthcare providers in a single interaction~\cite{meyer2021patient}. As a result, patients often exhibit a strong dependence on doctors for guidance and clarification. Doctors assume a leading role in the complex, multi-turn interactions with patients to gather the information required for accurate diagnosis and treatment~\cite{zhong2022hierarchical,chen2023dxformer}. Despite the crucial importance of this \emph{dynamic, patient-centered diagnostic process}, there is a scarcity of research assessing the ability of large language models (LLMs) to simulate this interaction. 

\begin{figure*}
    \centering
    \includegraphics[width=1.0\linewidth]{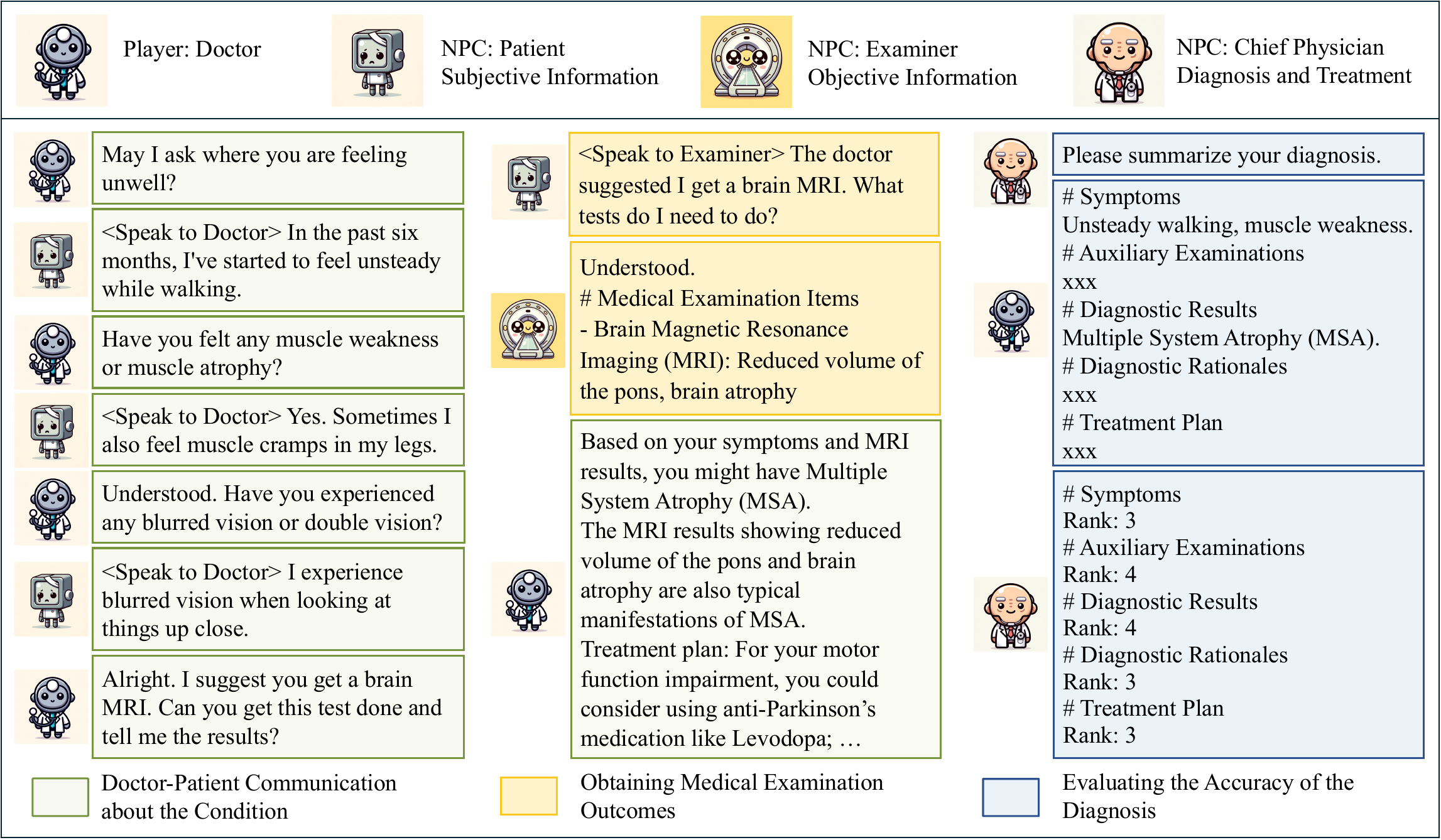}
    \caption{The demonstration of AI Hospital framework.}
    \label{fig:stepup_ai_hospital}
\end{figure*}

To address above challenges, we introduce \textbf{AI Hospital}, a LLM-powered multi-agent framework that simulates real-world dynamic medical interactions. AI Hospital consists of multiple non-player characters (NPCs), including \emph{Patient}, \emph{Examiner}, and \emph{Chief Physician}, as well as the player character, represented by the \emph{Doctor}. It recreates the scenario of a \emph{Patient} visit, requiring \emph{Doctor} to engage in multi-turn conversations with \emph{Patient}, ask relevant and probing questions, recommend appropriate medical examinations, and make diagnoses after collecting enough information. We set up \emph{Examiner} in the NPCs, who are specifically tasked with interacting with \emph{Patient} and providing pertinent medical examination results, ensuring that \emph{Doctor}  have access to the necessary objective information of the \emph{Patient} to make accurate diagnoses. Additionally, \emph{Chief Physician} is responsible for evaluating the performance of \emph{Doctor} after the entire session. The multi-agent nature of the AI Hospital framework allows for realistic simulations of complex medical scenarios, enabling a comprehensive assessment of an LLMs' ability to navigate various clinical situations. 

Based on the AI Hospital framework, we investigate the feasibility of utilizing LLMs as \emph{Doctor}s for clinical diagnosis by establishing the \textbf{Multi-View Medical Evaluation (MVME)} benchmark. This benchmark incorporates a collection of high-quality Chinese medical records meticulously screened by experienced medical professionals. These real-world cases provide detailed structured medical profiles, including patient subjective conditions, objective medical test results, diagnoses and treatments. Leveraging GPT-3.5 and GPT-4 to simulate the AI Hospital's non-player characters (NPCs), we conduct a thorough evaluation of the performance of LLMs-driven \emph{Doctor}s within this dynamic and realistic medical interaction environment. The MVME benchmark evaluates the performance of \emph{Doctor} along three key dimensions by \emph{Chief Physician}: the ability to \emph{collect symptoms}, \emph{recommend examinations}, and \emph{make diagnoses}. As a supplement, we also develop a link-based approach that integrates medical standard knowledge to generate \emph{entity-level} evaluation metrics. 

To enhance LLMs' diagnostic accuracy, drawing on previous research~\cite{croft2015science,centor2019pursuit} that highlights the importance of teamwork in clinical diagnosis, we delve into the collaborative mechanism~\cite{o2010teamwork,lamb2011teamwork}, facilitating iterative discussions among doctors. We utilize multiple \emph{Doctor}s to independently engage with the same medical record, allowing for diverse conversation trajectories and diagnostic outcomes. A dispute resolution strategy is proposed that effectively engages a \emph{Centre Agent} to guide discussions, clarify issues, and steer the \emph{Doctors} towards a more structured and efficient collaboration, thereby fostering more focused discussions and accelerating the achievement of consensus. 

To assess the performance of LLMs in clinical diagnosis, we conduct extensive experiments within the AI Hospital framework. We first validate the reliability of the various roles in AI Hospital and then evaluate a range of LLMs in the interactive diagnostic process. Our results reveal a \emph{substantial performance gap between interactive LLMs and the one-step GPT-4}, which serves as an upper bound by directly utilizing all patient information in a single interaction. On key metrics such as the accuracy of diagnostic results, diagnostic rationales, and treatment plans, the performance of interactive LLMs is less than 50\% of that achieved by the one-step GPT-4. Despite dedicated prompting engineering, LLMs struggle to make reasonable decisions in multi-turn interactions, leading to suboptimal diagnostic accuracy. The dispute resolution collaborative mechanism enhances performance to a certain degree but also falls short of the upper bound. It's suggested that existing LLMs may not have fully assimilated effective multi-turn diagnostic strategies. Our \textbf{quantitative results} highlights the challenges existing LLMs face in \textbf{posing pertinent questions, eliciting crucial symptoms, and recommending appropriate medical examinations}. These findings underscore the difficulties encountered by current LLMs in replicating the complex clinical reasoning processes employed by professional doctors and emphasize the need for further research to bridge the gap between LLMs and human physicians in clinical diagnosis.

In summary, the \textbf{main contributions} of this paper can be summarized as follows: 1) We introduce AI Hospital, a novel LLM-powered multi-agent framework to simulate medical interactions, enabling comprehensive evaluation of LLMs' ability to navigate complex clinical scenarios; 2) We establish the Multi-View Medical Evaluation (MVME) benchmark, which leverages high-quality medical records to evaluate the performance of LLMs-driven doctors in collecting symptoms, recommending examinations, and making diagnoses; 3) We propose a dispute resolution collaborative mechanism that facilitates iterative discussions among doctors to enhance diagnostic accuracy. The potential of our AI Hospital framework is comprehensively discussed in Appendix~\ref{sec:potential}.

\section{Setup of AI Hospital}
As depicted in Figure~\ref{fig:stepup_ai_hospital}, the AI Hospital framework comprises three \textbf{NPC characters} — the \underline{\emph{Patient}}, \underline{\emph{Examiner}}, and \underline{\emph{Chief Physician}} — and one \textbf{player character}, the \underline{\emph{Doctor}}. Each character assumes specific roles and responsibilities within the framework. The AI Hospital operates in two phases. In the \textbf{diagnostic phase}, \emph{Patient}, \emph{Examiner}, and \emph{Doctor} engage in conversations to exchange information necessary for accurate diagnosis. The number of interaction turns in this phase can vary depending on the \emph{Doctor}'s diagnostic strategy. Subsequently, during the \textbf{evaluation phase}, \emph{Chief Physician} is responsible for scoring the performance of \emph{Doctor} in the diagnostic phase. The following sections will elaborate on the settings and construction methods for each agent in AI Hospital.

\subsection{Agents Setup with Medical Records}
\label{section_medical_records}

Medical records are a valuable resource for reconstructing the hospital visit experience and simulating real-world medical interactions. By leveraging these records, we can reverse-engineer the diagnostic process and shape the behavior of agents within the AI Hospital framework. We categorize the information in each medical record into three types: 1) \textbf{Subjective Information} \quad This category includes the patient's symptoms, etiology, past medical history, habits, etc., which are primarily provided by the patient during their verbal interactions with the doctor; 2) \textbf{Objective Information} \quad This category encompasses medical test reports such as complete blood counts, urinalysis, and chest X-rays. The presence of these data in medical records indicates that the patient underwent these tests during the diagnostic process at the doctor's recommendation; 3) \textbf{Diagnosis and Treatment} \quad This category consists of \emph{diagnostic results}, \emph{diagnostic rationales}, and \emph{treatment courses}, which are the final conclusions made by the doctor during the diagnostic process, based on the combination of subjective and objective information.

These categories of information are assigned to the corresponding agents in the AI Hospital framework. \emph{Patient} has access to the subjective information, \emph{Examiner} is aware of the objective information, and \emph{Chief Physician} possess all information, while \emph{Doctor} do not have access to any information. AI Hospital framework assigns specific categories of information from medical records to each agent, shaping their scope of information within the diagnostic process.

\subsection{Agent Behavior Setting for NPCs}
\label{sec:agent_behavior}

In the AI Hospital framework, we leverage GPT-3.5 to power \emph{Patient} and \emph{Examiner}, and GPT-4 to drive \emph{Chief Physician}, enabling them to embody their roles authentically. Beyond providing NPCs with relevant information in medical records, we also employ meticulous prompt engineering to encourage they exhibit realistic behavior patterns. 

\textbf{Patient} \quad The \emph{Patient} agent is designed to exhibit a set of realistic behavior patterns to enhance the authenticity of the medical simulation:  1) \textbf{Cooperation}. The agent should actively respond to the doctor's inquiries and provide truthful answers, even if they may not proactively disclose all relevant information. The agent should actively participate in medical examinations recommended by the doctor; (2) \textbf{Communication}. The agent should use colloquial language and may omit certain important details or have subjective biases in describing their condition due to limited medical knowledge or personal beliefs; 3) \textbf{Curiosity}. The agent should express concerns or questions based on their level of understanding, seeking clear explanations from the doctor to address their doubts about the diagnosis or treatment process; 4) \textbf{Personalisation}. For each medical record, we employ GPT-4 to reason and imagine the patient’s unique background, experiences, emotional responses, and personality traits, thereby enhancing the realism and depth of the simulation. The prompt for \emph{Patient} agent is shown in Table~\ref{prompt:patient}.

\textbf{Examiner} \quad The \emph{Examiner} agent's primary goal is to provide relevant examination results when the Patient agent requests a query for a specific medical test. To maintain the authenticity of the simulation, the agent follows a realistic workflow. Upon receiving an examination query, the agent first identifies the requested medical examination and rejects any ambiguous or unclear requests. If the corresponding medical examination results are available, the Examiner agent returns the relevant findings to the doctor. In cases where no specific results are found, the agent reports no abnormalities. The Prompts are shown in Table~\ref{agent:examiner:step2} and ~\ref{agent:examiner:step1}. 

\textbf{Chief Physician} \quad The primary responsibility of the \emph{Chief Physician} agent is to evaluate the performance of the \emph{Doctor} agent in interactive diagnosis. After the diagnostic phase, \emph{Chief Physician} first requires the \emph{Doctor} to provide a comprehensive summary report for the patient, and then evaluates the summary report by comparing it with raw medical record, which serves as the gold standard. The prompts for \emph{Chief Physician} agent and more detailed description of the evaluation process can be found in Table~\ref{agent:evaluate:zh} and \S~\ref{section_collaboration} respectively.

\subsection{Agent Behavior Setting for Player}

The player agent, i.e., the \emph{Doctor}, can be powered by various LLMs that are being evaluated. However, in order to be able to engage in conversations based on predefined settings, LLMs are required to be well instruction-followed, otherwise LLMs will struggle to interact in AI Hospital.

\textbf{Doctor} \quad The \emph{Doctor} agent is designed to emulate the essential qualities and duties of a skilled and empathetic physician in real-world practice. The agent is encouraged to actively gather information, focusing on obtaining the patient's physical conditions like symptoms and medical history. When the agent determines that additional objective data is necessary to arrive at a confident diagnosis or to confirm a suspected condition, it suggests relevant examinations and tests. By synthesizing both subjective and objective findings, the agent aims to accurately diagnose the patient's condition, mirroring the systematic approach used by experienced doctors. The prompt of intern doctor is shown in Table~\ref{prompt:test_doctor}.

\subsection{Dialogue Flow in AI Hospital}
\label{dialogue_flow}

The AI Hospital framework simulates a realistic diagnostic process through a structured dialogue flow involving multiple agents. The conversation is initiated by the \emph{Patient} agent, who presents a chief complaint generated by GPT-4 based on the patient's medical record. The \emph{Doctor} agent then engages in a series of interactions with the \emph{Patient} and \emph{Examiner} agents to gather necessary information and make an accurate diagnosis. Throughout the dialogue, each agent's responses are prefixed with special symbols to explicitly indicate the intended recipient of their message, enabling a seamless multi-party conversation flow. The dialogue continues until the \emph{Doctor} agent reaches a diagnosis or a predefined maximum number of interaction rounds is reached. For a more detailed description of the dialogue flow, please refer to Appendix~\ref{appendix:dialogue_flow}. 

\section{MVME: Evaluation of LLMs as Intern Doctors for Clinical Diagnosis}
\label{section_collaboration}

Based on AI Hospital, we assess the feasibility of employing LLMs as \emph{Doctor} agent for clinical diagnosis by establishing the Multi-View Medical Evaluation (MVME) benchmark. 

\subsection{Multi-View Evaluation Criteria}
\label{sec:mvme}

Evaluating the performance of the \emph{Doctor} agent is a crucial component of the AI Hospital framework. As mentioned in \S~\ref{sec:agent_behavior}, in the evaluation phase, the \emph{Doctor} is required to provide a comprehensive summary report of the \emph{Patient}. We require the summary report consists of 5 parts, including the patient's \emph{symptoms}, \emph{medical examinations}, \emph{diagnostic results}, \emph{diagnostic rationales}, and \emph{treatment plan}.

Since the contents of the patient's medical record are described using natural language, the Chief Physician, as the evaluator, will directly compare each part of the report with the patient's complete medical record. For each part of the summary report, the GPT-4-driven Chief Physician needs to score from four discrete scores: 1, 2, 3, and 4, representing poorest to excellent performance. The evaluation of the "symptoms" part can reflect the comprehensiveness of the symptoms collected by the \emph{Doctor} during the interaction process. The evaluation of the "medical examinations results" part can reflect the appropriateness of the medical examinations suggested by the \emph{Doctor}. The evaluation of the other parts can reflect the \emph{Doctor}'s diagnostic and treatment capabilities. These metrics can reflect the LLMs' both \textbf{dynamic and static medical decision-making abilities}, including proactive inquiry, information gathering, clinical knowledge and comprehensive judgment.

In addition to the above model-based evaluation method, we also compute entity-overlap-based automated metrics for the diagnostic results part. We extract all disease entities from the diagnostic results provided by the LLMs and the actual medical records, and link them to their corresponding standardized disease entities. We then calculate the entity overlap to measure the accuracy of the final diagnoses made by the LLMs. We report the average number of extracted disease entities (\#), set-level precision (P), recall (R), and F1 score (F) metrics. Currently, we only calculate entity-level metrics for diseases because readily available entity linking methods and standards, such as the International Classification of Diseases (ICD-10)~\cite{trott1977international}, exist for this purpose. For symptoms and examinations, linking them to corresponding entities is more challenging due to the lack of readily available tools. Therefore, we do not incorporate their entity-level metrics in this paper.

\subsection{MVME Dataset Construction}
\label{sec:data_preparation}

\begin{wraptable}{r}{0.4\textwidth}
    \centering
    \caption{Departments distribution.}
    \label{table:department_distribution}
    \begin{tabular}{cc}
    \toprule
    Department & \# \\
    \midrule
    Surgery & 180 \\
    Internal Medicine & 153 \\
    Obstetrics and Gynecology &94  \\
    Pediatrics &29  \\
    Otorhinolaryngology &23  \\
    Others & 27 \\
    \bottomrule
\end{tabular}
\end{wraptable}

We collect Chinese medical records across diverse departments online \footnote{\href{https://bingli.iiyi.com/}{https://bingli.iiyi.com/}} and engage professional physicians for a thorough review. Subsequent to the exclusion of records with deficiencies, such as incomplete information, a total of 506 cases remained. The detailed distribution of these cases among the various departments is presented in Table~\ref{table:department_distribution}. 

To verify the quality of the collected medical records, we select samples from 10 secondary departments, randomly choosing five cases per department for review. Doctors from corresponding departments are hired to evaluate the ``Diagnosis and Treatment'' including diagnostic results, diagnostic rationales, and treatment course with a binary choice: either ``fundamentally correct'' or ``obviously incorrect''. If three sections of a medical record are fundamentally accurate, then we consider the medical record to be correct. Overall, expert validation conclude that 94\% of the records are deemed as correct.

\section{Collaborative Diagnosis of LLMs Focused on Dispute Resolution}
\label{sec:collaboration}

To further improve diagnostic accuracy, we propose a collaborative mechanism for clinical diagnosis that leverages the power of multiple LLMs. In our collaborative framework, we employ different LLMs to serve as individual \emph{Doctors}, each engaging in interactive consultations with the \emph{Patient}. Due to the inherent differences among LLMs, these interactions may result in diverse dialogue trajectories and diagnostic reports. To streamline the process of forming a unified diagnostic report, we introduce a \emph{Central Agent}, also referred to as the \emph{Chief Physician}, to participate as a moderator. The overall process is shown in Figure~\ref{fig:collborative_diagnosis}. 

The \emph{Chief Physician} consolidates and analyzes the data collected from \emph{Doctors}, confirms disputed points with \emph{Patient} and \emph{Examiner}, and synthesizes a comprehensive summary of the patient's condition. Through multiple discussion iterations, the \emph{Chief Physician} identifies key points of disagreement among \emph{Doctors} and guides them to engage in targeted discussions, progressively refining their understanding and working towards a consensus. This collaborative mechanism harnesses the collective intelligence of LLMs to enhance the accuracy and robustness of clinical diagnosis by capitalizing on their diverse knowledge and reasoning capabilities while promoting a structured and iterative process of refining diagnostic reports. The overall process is delineated as pseudocode in Algorithm~\ref{algorithm:1}, and the prompts are listed in Table~\ref{agent:medical_director} and Table~\ref{agent:doctor} in the appendix.

\section{Experiments}
\subsection{Agent Behavior Analysis in AI Hospital Framework}

Before presenting the main results, it is crucial to verify whether the agents in the AI Hospital framework effectively align with their intended roles and behaviors. We conduct a experiment to investigate the behaviors of several key agents, including the \emph{Patient}, \emph{Examiner}, and \emph{Doctor}. 

\begin{table}[!th]
    \caption{Human evaluation for agent behavior in AI Hospital. \# represents the sample size, such as number of total doctor-patient QA pairs in 50 dialogues.} 
    \label{table:human_eval_p1}
    \vspace{2mm}
    \centering
    \begin{tabular}{l|ccc|cc|c}
        \toprule
        &\multicolumn{3}{c}{Patient} &\multicolumn{2}{c}{Examiner} &Doctor \\
          & \#  &Relevance &Honesty & \#  &Accuracy &Consistency \\
        \midrule
        Qwen-Max &429 &100.0\% &99.0\% &56 &98.2\% &99.0 \\
        Wenxin-4.0 &472 &100.0\% &98.1\% &68 &98.5\% &99.0 \\
        GPT-3.5 &417 &100.0\% &99.5\% &57 &98.2\% &98.0 \\
        GPT-4 &378 &100.0\% &99.7\% &61 &100.0\% &100.0 \\
        \bottomrule
    \end{tabular}
\end{table}

\textbf{Evaluation Metric} \quad For the Patient agent, we focus on two dimensions in the communication between the patient and the doctor. The first dimension is the \textbf{relevance} of the patient's responses to the doctor's questions. The second dimension is the \textbf{honesty} of the patient's responses with the subjective information in the medical record. For the Examiner agent, we assess the \textbf{accuracy} of the agent's understanding of the requested medical examination and its ability to return the corresponding examination results when receiving a query for a medical examination. For the Doctor agent, we evaluate the \textbf{consistency} of the doctor's final diagnostic report with the information in the dialogue flow. We categorize the consistency into three levels: 1) significantly inconsistent, 2) slightly inconsistent, and 3) mostly consistent. These levels are assigned scores of 1, 2, and 3, respectively. Finally, we map this score to a range of 0-100. We document our evaluation methodology in detail in Appendix~\ref{human_eval_agent_behavior}.

\textbf{Experimental Setup} \quad We employ multiple \emph{Doctor} agents, including GPT-3.5, GPT-4~\cite{OpenAI2023GPT4TR}, Wenxin-4.0, and Qwen-Max~\cite{bai2023qwen}. We randomly select 50 medical record samples and ask each agent generate 50 multi-turn dialogue trajectories within the AI Hospital framework. We manually label all the metrics and report the average values. 

\textbf{Results and Analysis} \quad Table~\ref{table:human_eval_p1} demonstrates the effectiveness of the AI Hospital framework in simulating realistic medical interactions, with high scores (all over 95) across all metrics indicating reliable and consistent agent behaviors. The \emph{Patient} agent can provide accurate and pertinent information, the \emph{Examiner} agent can accurately understand and return requested medical examination results, and the \emph{Doctor} agent can generate consistent diagnostic reports. It validates the reliability and effectiveness of the proposed multi-agent system, laying a solid foundation for assessing LLMs' performance in clinical diagnosis. 

\subsection{Can LLMs Diagnose Like Doctors?}

In this section, we investigate the core question of this paper, i.e., can LLMs make diagnoses like doctors?  Based on the AI Hospital, We evaluate a range of LLMs, including GPT~\cite{OpenAI2023GPT4TR} (GPT-3.5 and GPT-4), Wenxin-4.0, QWen-Max~\cite{bai2023qwen}, Baichuan 13B, HuatuoGPT-II 13B and 34B~\cite{chen2023huatuogpt}. Among them, HuatuoGPT-II is designed specifically for the medical field. We only select HuatuoGPT-II as the comparative model specifically because many medical LLMs have significantly lost their instruction-following capabilities during the training process. This loss makes it difficult for these models to adhere to our prompts and engage in meaningful dialogue, resulting in poor performance on our benchmark.


\begin{table*}[t]
\caption{MVME: GPT-4 evaluation with reference in clinical consultation. GPT-4$^{*}$ in One-Step is the upper bound. For GPT-4$^{*}$, the ground truth of symptoms and medical examinations are provided, resulting in a score of 100.0.} 
\label{table:llm_evaluation}
\centering
\resizebox{1.0\textwidth}{!}{
\begin{tabular}{l|cccccc}
\toprule
&\multirow{2}{*}{{Symptoms}} &{Medical} &{Diagnostic} &{Diagnostic} &{Treatment} \\
& &{Examinations} &{Results} &{Rationales} &{Plan} \\
\midrule
&\multicolumn{5}{c}{Interaction} \\
\midrule
Baichuan (13B) &52.56\ (2.77) &21.06\ (2.83) &19.50\ (2.74) &17.40\ (2.51) &13.97\ (2.37) \\
HuatuoGPT-II (13B) &61.06\ (2.17) &24.43\ (2.50) &20.03\ (2.56) &20.03\ (2.37) &14.23\ (2.18) \\
HuatuoGPT-II (34B) &68.43\ (1.83) &30.30\ (2.77) &25.20\ (2.52) &27.46\ (2.55) &21.33\ (2.37) \\
GPT-3.5 &66.39\ (1.33) &\textbf{31.03}\ (2.97) &23.90\ (2.43) &24.43\ (2.42) &17.73\ (2.17) \\
Wenxin-4.0 &67.79\ (1.33) &30.43\ (2.70) &26.23\ (2.63) &26.46\ (2.57) &22.00\ (2.43) \\
Qwen-Max &61.69\ (2.10) &26.60\ (2.63) &26.46\ (2.63) &28.76\ (2.63) &24.90\ (2.45) \\
GPT-4 &\textbf{69.03}\ (1.27) &25.10\ (2.63) &\textbf{29.36}\ (2.58) &\textbf{30.76}\ (2.57) &\textbf{26.93}\ (2.63) \\
\midrule
&\multicolumn{5}{c}{Collaboration} \\
\midrule
2 Doctors w/o DR &75.49\ (2.03) &43.03\ (3.03) &35.56\ (2.83) &38.53\ (2.76) &32.40\ (2.60)  \\
2 Doctors &78.06\ (1.83) &47.10\ (2.97) &38.06\ (2.72) &41.56\ (2.75) &35.90\ (2.62) \\
3 Doctors &\textbf{80.26}\ (1.80) &\textbf{49.63}\ (2.83) &\textbf{39.60}\ (2.80)  &\textbf{44.23}\ (2.77) &\textbf{37.26}\ (2.63) \\
\midrule
&\multicolumn{5}{c}{One-Step} \\
\midrule
GPT-4$^{*}$  &100.0$^{*}$ &100.0$^{*}$ &$58.89\ (1.63)$ &$66.59\ (1.33)$ &$53.16\ (1.83)$\\
\bottomrule
\end{tabular}
}
\end{table*}

\textbf{Evaluation} \quad As mentioned in \S~\ref{sec:mvme}, we employ the proposed multi-view evaluation criteria. We normalize the scores of all metrics to a range between 0 and 100 and utilize the classic bootstrap method~\cite{efron1992bootstrap} to compute the variance. 

\begin{wraptable}{r}{0.50\textwidth}
    \centering
    \caption{MVME: Link-based evaluation of diagnostic results.}
    \label{table:link_evaluation}
    \vspace{2mm}
    \resizebox{0.5\textwidth}{!}{
    \begin{tabular}{l|ccccc}
    \toprule
    &{\#} &{R}  &{P} &{F1} \\ 
    \midrule
    &\multicolumn{4}{c}{Interaction} \\
    \midrule
    Baichuan (13B) &1.58 &10.21 &23.79 &14.28 \\
    HuatuoGPT-II (13B) &1.72 &12.76 &24.84 &16.85 \\
    HuatuoGPT-II (34B) & 1.86 &17.48& 30.95 & 22.34 \\
    GPT-3.5 &1.81 &19.19& 37.39 & 25.37  \\
    Wenxin-4.0 &2.50 &22.03& 31.44 & 25.91 \\
    Qwen-Max & 1.77 &\textbf{22.42} & 43.38 & 29.56 \\ 
    GPT-4 &1.52 &21.64&\textbf{50.26} & \textbf{30.26} \\
    \midrule
    &\multicolumn{4}{c}{Collaboration} \\
    \midrule
    2 Doctors w/o DR &2.37 &28.44 &41.45 &33.74 \\
    2 Doctors  &2.41 &29.51 &\textbf{43.62} &35.21 \\
    3 Doctors &3.20 &\textbf{36.54} &39.58 &\textbf{38.00} \\
    \midrule
    &\multicolumn{4}{c}{One-Step} \\
    \midrule
    GPT-4$^{*}$  &2.30 &38.90 &58.97 &46.88 \\
    \bottomrule
\end{tabular}
}
\end{wraptable}

\textbf{One-Step Diagnosis as Upper Bound} \quad In the one-step diagnosis, we directly feed the patient's subjective information and objective information described in \S~\ref{section_medical_records} as input to GPT-4, prompting it to generate a diagnostic report without going through the interactive diagnostic phase. We consider the performance of GPT-4 in this one-step setting as the upper bound of LLM performance.

\textbf{Interactive Diagnostic Performance of LLMs} \quad The main experimental results are presented in Table~\ref{table:llm_evaluation} and Table~\ref{table:link_evaluation}. Our findings reveal several key insights into the performance of existing LLMs in the AI Hospital framework. One of notable observations is that the diagnostic performance of existing LLMs in the AI Hospital framework falls significantly short of the upper bound set by the one-step GPT-4 approach. Even GPT-4 achieves less than half of the upper bound performance. This finding highlights the substantial limitations of current LLMs in interactive settings, suggesting that they have not yet learned sufficiently rich real-world clinical decision-making experiences. We also observe that LLMs with less parameters tend to exhibit weaker interactive abilities, such as Baichuan (13B), demonstrates lower performance in interactive diagnosis.

\textbf{Gather Information Helps Diagnose} Based on Table~\ref{table:llm_evaluation}, we further explore the relationship between the information finally collected and the quality of diagnosis. We use Symptoms and Medical Examinations to measure the completeness of patient information, and use Diagnostic Results, Diagnostic Rationales, and Treatment Plans to evaluate diagnostic quality. By using linear regression, we present our results in Figure~\ref{fig:lr}, which show that there is a significant positive correlation between more complete patient information and higher diagnostic quality. This further explains the shortcomings of current LLMs, namely that it is difficult for LLMs to collect patients' symptoms through active questioning like doctors, and it is even more difficult for them to recommend correct medical examinations. \textbf{This lack of dynamic clinical decision-making ability is a huge obstacle that prevents LLMs from diagnosing like doctors}. The details of Figure~\ref{fig:lr} can be found in Appendix~\ref{sec:lr_mvme}.

\section{Further Analysis}
\subsection{Collaboration Mechanism}

In Table~\ref{table:llm_evaluation}, we also evaluate several models with different settings of the cooperation mechanism. The comparative methods include Collaborative Diagnosis with 3 and 2 Agents, an 2 Agents without Dispute Resolution. They are denoted as 3 Doctors, 2 Doctors and 2 Doctors w/o DR. The initial two intern doctors are powered by GPT-3.5 and GPT-4 for interactive consultation, while the last one is using Wenxin-4.0.

\begin{figure}[!th]
    \centering
    \includegraphics[width=0.55\linewidth]{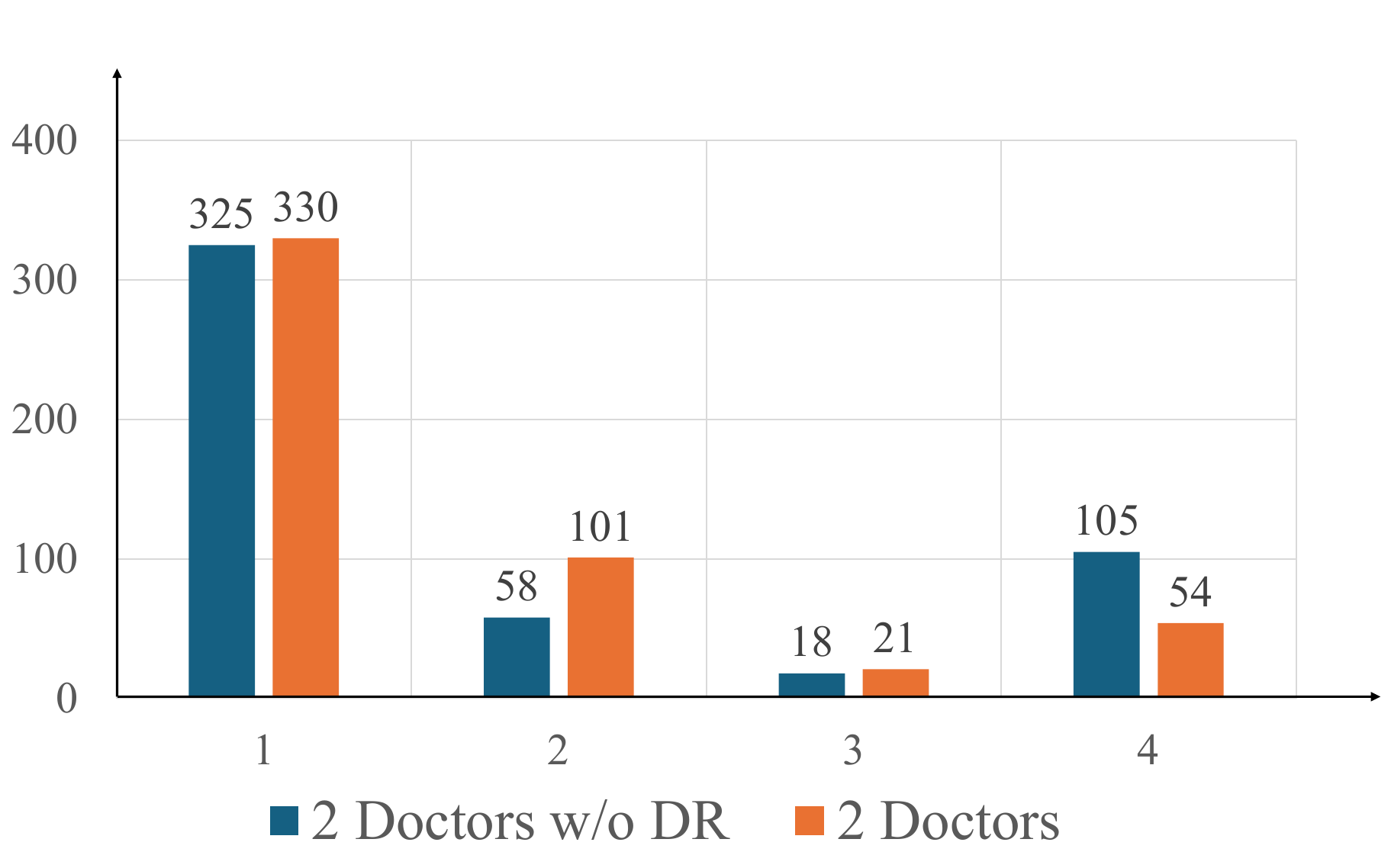}
    \caption{Statistical analysis of discussion rounds in collaborative frameworks with and without ``Dispute Resolution'' mechanism.}
    \label{fig:discussion-turn}
\end{figure}

\textbf{Effectiveness Collaboration Mechanism} \quad We observed several key findings: 1) The collaborative use of models can exceed the performance of GPT-4, thereby validating the efficacy of the cooperative mechanism; 2) Collaboration among ``3 Doctors'' enhances diagnosis compared to ``2 Doctors'', highlighting the benefits of more agents in cooperation; 3) The removal of the ``Dispute Resolution'' mechanism from the ``2 Doctors'' reduces its effectiveness, emphasizing the significance of establishing a better consensus.

\textbf{Efficiency of Dispute Resolution in Collaboration} \quad For the ``Dispute Resolution'', we continue to check whether intern doctors can reach consensus more rapidly. In terms of efficiency, a comparative analysis is conducted on the number of discussion rounds necessary to achieve consensus, both with and without the ``Dispute Resolution'' mechanism. The outcomes are detailed in Figure~\ref{fig:discussion-turn}. These findings reveal a marked increase in the rate of consensus achieved within the initial four discussion rounds following the adoption of the dispute resolution mechanism. This enhancement suggests that the process, facilitated by the \emph{Chief Physician} highlighting controversial issues and \emph{Doctor}s concentrating on these discussions, effectively reduces the time required to achieve consensus.

\subsection{Reasons for Failure Cases}

We analyze an analysis on 219 cases where GPT-4 render incorrect diagnostic results, and rated as 1 point by the \emph{Chief Physician}. Through a systematic manual review, these errors are mainly categorized into three distinct types, which are detailed in Table~\ref{table:error_types}. 

\begin{wraptable}{r}{0.5\textwidth}
    \centering
    \caption{Classification and statistics of misdiagnoses (1 point) of the intern doctor powered by GPT-4.}
    \label{table:error_types}
    \begin{tabular}{cc}
        \toprule
        Error Type & \# \\
        \midrule
        Omission of auxiliary examinations & 99 \\
        Exclusive focus on complications & 52 \\
        Erroneous judgment & 68 \\
        \bottomrule
    \end{tabular}
\end{wraptable}

\textbf{Omission of Auxiliary Examinations} \quad An illustrative case involved the failure to detect gallbladder stones, attributed to the absence of a recommended abdominal ultrasound. This category highlights instances where GPT-4 did not suggest essential auxiliary examinations that could have potentially confirmed or ruled out possible medical conditions.

\textbf{Exclusive Focus on Complications} \quad In certain cases, GPT-4 focuses only the symptoms given by the patient, such as soft tissue swelling in the feet, while ignoring underlying complications, such as diabetes. This type of error arises from the LLMs' limited recognition of the interconnectedness between symptoms and underlying health issues, and its failure to prompt further inquiry into the patient's comprehensive health status.

\textbf{Erroneous Judgment} \quad Even when presented with complete symptomatology and medical examination results, GPT-4 occasionally reach incorrect conclusions. This category of error points to a lack of sufficient medical expertise embedded within the LLMs, leading to diagnostic inaccuracies even with comprehensive data.

\section{Related Works}
\textbf{LLM Powered Agents} \quad Before the popularity of LLMs, there are already efforts to create agents in the medical field, particularly for medical education~\cite{watts2021healthcare,antel2022use}. However, these agents often lack flexibility, relying on rule-based or traditional machine learning algorithms made it difficult to accurately simulate the complexity of medical scenarios. The advancement of LLMs powered agents has led to significant strides in complex task resolution through human-like actions, such as tool-learning~\cite{chen2023disc,schick2024toolformer}, retrieval augmentation~\cite{yue2023disc,asai2023self}, role-playing~\cite{park2023generative}, communication~\cite{xi2023rise,wang2023survey}. This includes applications in software design and molecular dynamics simulation. Recent research~\cite{tang2023medagents} in the medical field has highlighted the critical roles and decision-making processes in medical QA, encompassing various investigations like CT scans, ultrasounds, electrocardiograms, and blood tests. Despite these advancements, effectively integrating LLM-based agents into the medical domain, particularly in disease diagnosis, presents a notable challenge~\cite{zhou2023survey}. Our research pioneers the use of multi-agent systems in creating a clinical diagnosis environment. We introduce a novel mechanism for identifying, discussing, and resolving disputes in collaboration, demonstrating promising results in clinical diagnosis. 

\textbf{Medical Large Language Models} \quad Prior to the emergence of large language models (LLMs), the majority of automated diagnostic methods~\cite{zhong2022hierarchical,chen2023dxformer} relies on reinforcement learning to guide agents in gathering symptoms and conducting diagnoses. The development of LLMs in the medical domain has been driven by open-source Chinese LLMs and various fine-tuning methods. Models like Med-PaLM~\cite{singhal2022large}, DoctorGLM~\cite{xiong2023doctorglm}, BenTsao~\cite{wang2023huatuo}, ChatGLM-Med~\cite{ChatGLM-Med}, Bianque-2~\cite{chen2023bianque}, ChatMed-Consult~\cite{zhu2023ChatMed}, MedicalGPT~\cite{MedicalGPT}, and DISC-MedLLM~\cite{bao2023discmedllm} fine-tune using different datasets, techniques, and frameworks, focusing on medical question answering, health inquiries and doctor-patient dialogues. 

\textbf{Evaluation in Medicine AI} \quad Prior research in medical AI evaluation has concentrated on non-interactive tasks, including question answering, entity and relation extraction, and medical summarization and generation. In biomedical question answering, key datasets such as MedQA (USMLE)~\cite{jin2021disease}, PubMedQA~\cite{jin2019pubmedqa}, and MedMCQA~\cite{pal2022medmcqa} are utilized, with accuracy serving as the primary evaluation metric. The objective of entity and relation extraction~\cite{li2020survey} is to categorize named entities and their relationships from unstructured text into specific predefined classes. Prominent biomedical NER datasets include NCBI Disease~\cite{dougan2014ncbi}, JNLPBA~\cite{collier2004introduction}, BC5CDR~\cite{li2016biocreative}, BioRED~\cite{luo2022biored} and IMCS-21~\cite{chen2023benchmark,chen2023knse}, with the F1 score being the standard for model performance assessment. Medical summarization and generation tasks involve converting structured data, like tables, into descriptive text. This includes the creation of patient clinic letters, radiology reports, and medical notes~\cite{liu2023utility}. The principal datasets for these tasks are PubMed~\cite{jin2019pubmedqa} and MentSum~\cite{sotudeh2022mentsum}. A recent study introduced BioLeaflets~\cite{yermakov2021biomedical} and assessed multiple Large Language Models (LLMs) in data-to-text generation.

\section{Conclusion}
In AI Hospital, we take a step forward in the field of medical interactions by concentrating on clinical diagnosis. We introduce AI Hospital to build a real-time interactive consultation scenario. We generate simulated patients and examiner using collected medical records and established a comprehensive engineering process. Based on the platform, we build a benchmark MVME to explore the feasibility of  different LLMs in interactive consultations. To improve the diagnostic accuracy, this research also introduces a novel collaborative mechanism for intern doctors, featuring iterative discussions and a dispute resolution process, supervised by a medical director. In our experiment, the results not only demonstrate the performance of different LLMs but also confirm the efficacy of our dispute resolution-centered collaborative approach. For in-depth analysis, we list the error types and identify the issues that should be addressed.  In the future, we will focus on building more comprehensive benchmark, cost-effective evaluation framework, and optimizing the agents.

\section*{Limitations}

The AI Hospital framework and MVME benchmark, while making significant strides in evaluating the interactive performance of LLMs in clinical diagnosis, have several limitations. The use of primarily Chinese medical records may limit generalizability to other languages and healthcare systems. Although diverse, the sample size of 506 cases may not fully capture the complexity of real-world scenarios, including rare diseases. Simulated interactions between agents may not perfectly replicate human-to-human nuances, requiring further validation. The current treatment plan evaluation system is insufficient, as it does not consider feasible alternative strategies, potentially underestimating LLMs' performance. Lastly, the extensive use of OpenAI's LLM API may increase the environmental burden, which could be mitigated by leveraging smaller, more efficient open-source models in future studies. Despite these limitations, the AI Hospital framework and MVME benchmark provide a solid foundation for future research on evaluating and improving LLMs' clinical diagnostic capabilities.

\section*{Ethics Consideration}

Ethical considerations are of utmost importance in our research on the application of LLMs in clinical diagnosis. We recognize the potential implications of our work and have taken steps to address them. Firstly, to ensure transparency and reproducibility, we will release the publicly accessible online medical records data used in our study. This allows other researchers to validate and build upon our findings, promoting collaborative progress in this field. However, we acknowledge the critical importance of privacy protection. The data sources have undergone a process of de-identification, removing sensitive information before our collection. Furthermore, we recognize the potential for bias in AI systems, which could perpetuate or amplify disparities in healthcare. To mitigate this risk, we have made efforts to ensure the diversity and representativeness of our medical record datasets. . By proactively addressing these considerations, we aim to realize the potential benefits of AI-assisted diagnosis while ensuring its responsible and equitable implementation.

\bibliographystyle{unsrt}
\bibliography{custom}

\appendix
\section*{Appendix}
\addcontentsline{toc}{section}{Appendix}
\section{Detailed Description of the Dialogue Flow in AI Hospital}
\label{appendix:dialogue_flow}

This section is the detailed description of the dialogue flow in AI Hospital in~\S\ref{dialogue_flow}. The AI Hospital framework aims to simulate a realistic diagnostic process by orchestrating a structured dialogue flow involving multiple agents, namely the \emph{Doctor}, \emph{Patient}, and \emph{Examiner}. This appendix provides a comprehensive description of the dialogue flow, detailing the initiation, progression, and termination phases.

\begin{figure*}[!th]
    \centering
    \includegraphics[width=1.0\textwidth]{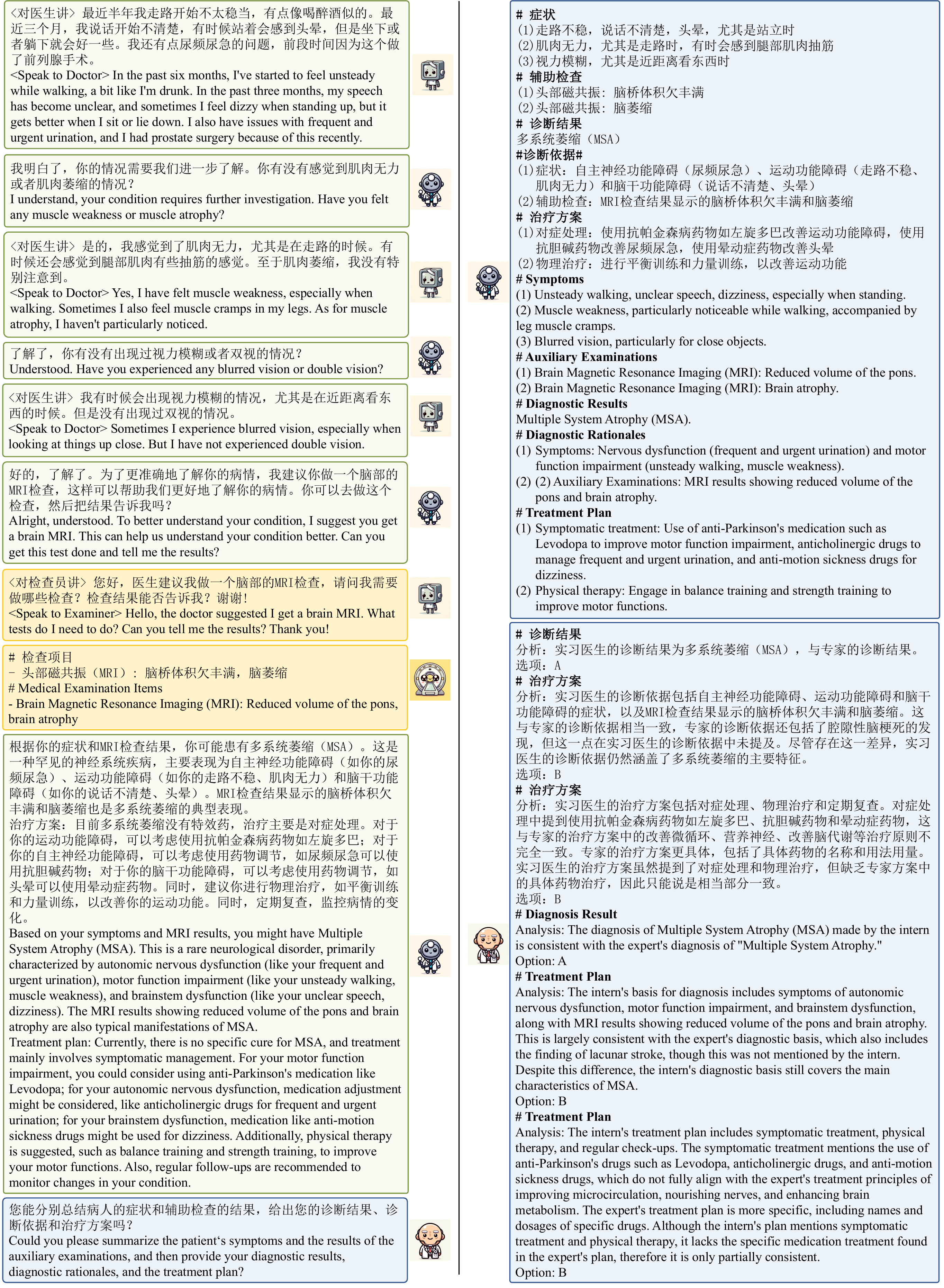}
    \caption{An example of dialogue flow among \emph{Doctor}, \emph{Patient}, \emph{Examiner} and \emph{Chief Physician} in AI Hospital framework.}
    \label{fig:case_study}
\end{figure*}

\paragraph{Dialogue Initiation} ~ {The dialogue commences with the \emph{Patient} agent generating a chief complaint based on the information contained in their medical record. To create this initial complaint, GPT-4 is employed to analyze the patient's medical record and generate a concise statement that encapsulates the patient's recent physical condition. The generated complaint is designed to align with the predefined persona of the patient, accurately reflecting their language style and focusing on a relevant subset of the subjective information available in the medical record. This chief complaint serves as the starting point for the first round of dialogue between the \emph{Patient} and \emph{Doctor} agents.}

\paragraph{Dialogue Progression} ~ {The diagnostic process unfolds through a series of interactions between the \emph{Doctor}, \emph{Patient}, and \emph{Examiner} agents. The \emph{Doctor} agent assumes an active role in this phase, engaging in a comprehensive inquiry to elicit detailed information about the patient's condition. This involves asking pertinent questions and recommending appropriate medical examinations to gather the necessary data for formulating an accurate diagnosis. The \emph{Patient} agent, serving as a non-player character (NPC), autonomously determines its course of action at each dialogue turn based on meticulously designed prompts. When communicating with the \emph{Doctor} agent, the \emph{Patient} agent prefaces its responses with the designated characters "<Speak to Doctor>". In these interactions, the \emph{Patient} agent provides answers to the doctor's inquiries and offers feedback on their physical condition. Conversely, when the \emph{Patient} agent needs to request examinations based on the doctor's instructions, it initiates communication with the \emph{Examiner} agent using the prefix "<Speak to Examiner>". The \emph{Patient} agent conveys the requested examination items to the \emph{Examiner} agent, who subsequently reports the corresponding examination results back to the \emph{Doctor} agent.}

\paragraph{Dialogue Termination} ~ {The termination conditions for the dialogue in the diagnostic phase are clearly defined within the \emph{Patient} agent's prompt. The dialogue reaches its conclusion when either of two conditions is satisfied. Firstly, if the \emph{Patient} agent receives the doctor's diagnostic results, it generates the special termination token "<END>", signaling the end of the diagnostic phase. Alternatively, the dialogue concludes when the predefined maximum number of interaction rounds is surpassed. These termination conditions ensure a structured and finite dialogue flow, preventing the diagnostic phase from continuing indefinitely. It is noteworthy that the number of rounds in the evaluation phase is predetermined, rendering termination conditions relevant only for the diagnostic phase.}

By adhering to this well-defined dialogue flow, the AI Hospital framework enables a systematic and realistic simulation of the diagnostic process, facilitating effective communication and information exchange among the \emph{Doctor}, \emph{Patient}, and \emph{Examiner} agents. This structured approach guarantees a coherent and logical progression of the dialogue, ultimately leading to a comprehensive evaluation of the \emph{Doctor} agent's performance. Figure~\ref{fig:case_study} shows a specific dialogue flow example in the AI Hospital framework. 

\begin{figure*}[b]
    \centering
    \includegraphics[width=1.0\linewidth]{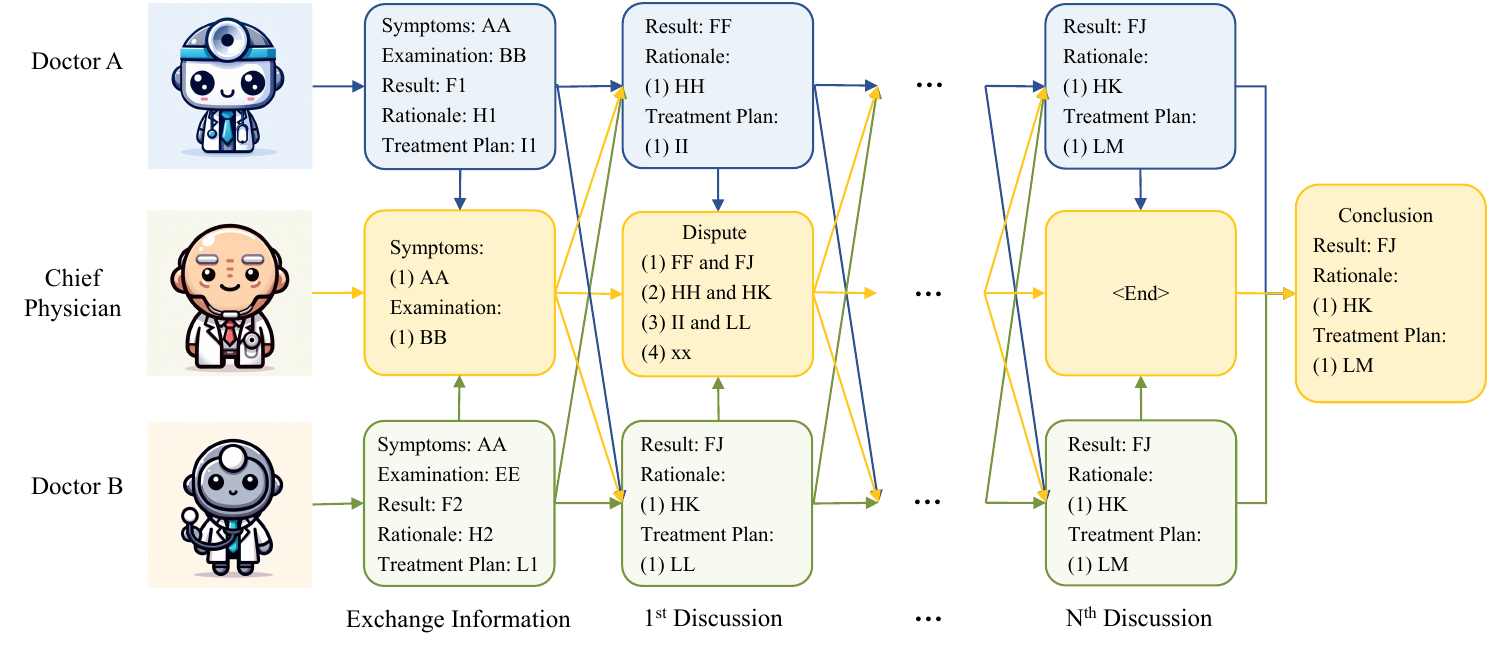}
    \caption{Collaboration of \emph{Doctors} for clinical diagnosis.}
    \label{fig:collborative_diagnosis}
\end{figure*}

\section{Detail of Collaborative Algorithm}

In this section, we delve into the details of our proposed multi-agent collaborative algorithm in \S~\ref{sec:collaboration}. In this process, the \emph{Central Agent}, also called \emph{Chief Physician}, does not exactly follow the behavior pattern of the \emph{Chief Physician} agent in AI Hospital framework mentioned in \S~\ref{section_medical_records}. The \emph{Central Agent} is unaware of any information, whose goal is to coordinate multiple \emph{Doctor} agents to collaboratively improve diagnosis. Figure~\ref{fig:lr} shows an example flow of the collaboration process, and the corresponding pseudocode is provided in Algorithm~\ref{algorithm:1}.

\subsection{Exchange of Factual Information}

We contend that a consensus on the physical condition of patients among \emph{Doctors} constitutes the cornerstone of collaborative diagnosis. The process of building the consensus is delineated into three distinct steps, outlined below. 

\begin{itemize}
    \itemsep 0pt
    \item uring the interactive consultation process, \emph{Doctor} agents communicate with the \emph{Central Agent}, i.e., \emph{Chief Physician}, relaying patient factual information they have acquired, focusing primarily on symptoms and medical test outcomes.
    \item The \emph{Chief Physician} consolidates and analyzes the data collected from multiple \emph{Doctor}s, confirming symptoms and test outcomes with \emph{Patient} and \emph{Examiner} to clarify disputed points.
    \item Drawing upon the findings received from \emph{Doctor}s, coupled with feedback from \emph{Patient} and \emph{Examiner}, the \emph{Chief Physician} synthesizes a comprehensive summary of the symptoms and medical examination outcomes.
\end{itemize}

\begin{algorithm}[th]
\caption{Dispute Resolution Collaboration}
\begin{algorithmic}[1]
\REQUIRE Maximum number of rounds $M$, number of intern doctors $N$ and pre-diagnosis $P$.
\ENSURE Final Diagnosis $a$
\STATE $D$ \COMMENT{Medical Director}
\STATE $I \gets [I_1,\cdots,I_N]$ \COMMENT{Intern Doctors}
\STATE $H \gets P$   \COMMENT{Initialize Discussion History}
\STATE $d \gets D(H)$  \COMMENT{Initialize Dispute}
\STATE $m \gets 0$     \COMMENT{Current Round}
\WHILE{$m \leq M$}
    \STATE $m \gets m + 1$
    \FOR{each $I_i$ in $I$}
        \STATE $h \gets D_i(H, d)$  \COMMENT{Generate Diagnosis}
        \STATE $H \gets H + [h]$ \COMMENT{Append $h$ to $H$}
    \ENDFOR
    \STATE $d \gets D(H)$       \COMMENT{Summarize Disputes}
    \IF{$d$ \emph{is NULL}}                
        \STATE break            \COMMENT{Debate is Over}
    \ENDIF
\ENDWHILE
\STATE $a \gets D(H)$
\end{algorithmic}
\label{algorithm:1}
\end{algorithm}

\subsection{Discussions on Dispute Resolution}

In collaborative diagnosis, the \emph{Chief Physician} should analyze the statements of \emph{Doctors} and identify key points of disagreement to foster focused discussions. The process is as follows: 

\begin{itemize}
    \itemsep 0pt
    \item The collaborative diagnosis consists of multiple discussion iterations. Under the guidance of the \emph{Chief Physician}, \emph{Doctors} are expected to delve deeper gradually, resolve differences, and reach a consensus.
    \item In each session of collaborative diagnostic discussion, each \emph{Doctor} should present their diagnostic reports while engaging in critical analysis of their peers' findings. Guided by the \emph{Chief Physician}'s summary of disputed points among \emph{Doctors}, they can pinpoint the current issues requiring attention. This approach facilitates targeted and thorough critical thinking of \emph{Doctors}, enhancing the refinement of their reports.
    \item Upon the conclusion of discussions, the \emph{Chief Physician} assesses the persistence of disagreements among interns. If disagreements are identified, the director can summarize the controversial issues and set then as the agenda for the subsequent session to facilitate resolution. Conversely, if no disagreements are found, the director concludes the discussions and finalizes the diagnostic report by himself.
\end{itemize}

\begin{figure}
    \centering
    \includegraphics[width=0.9\linewidth]{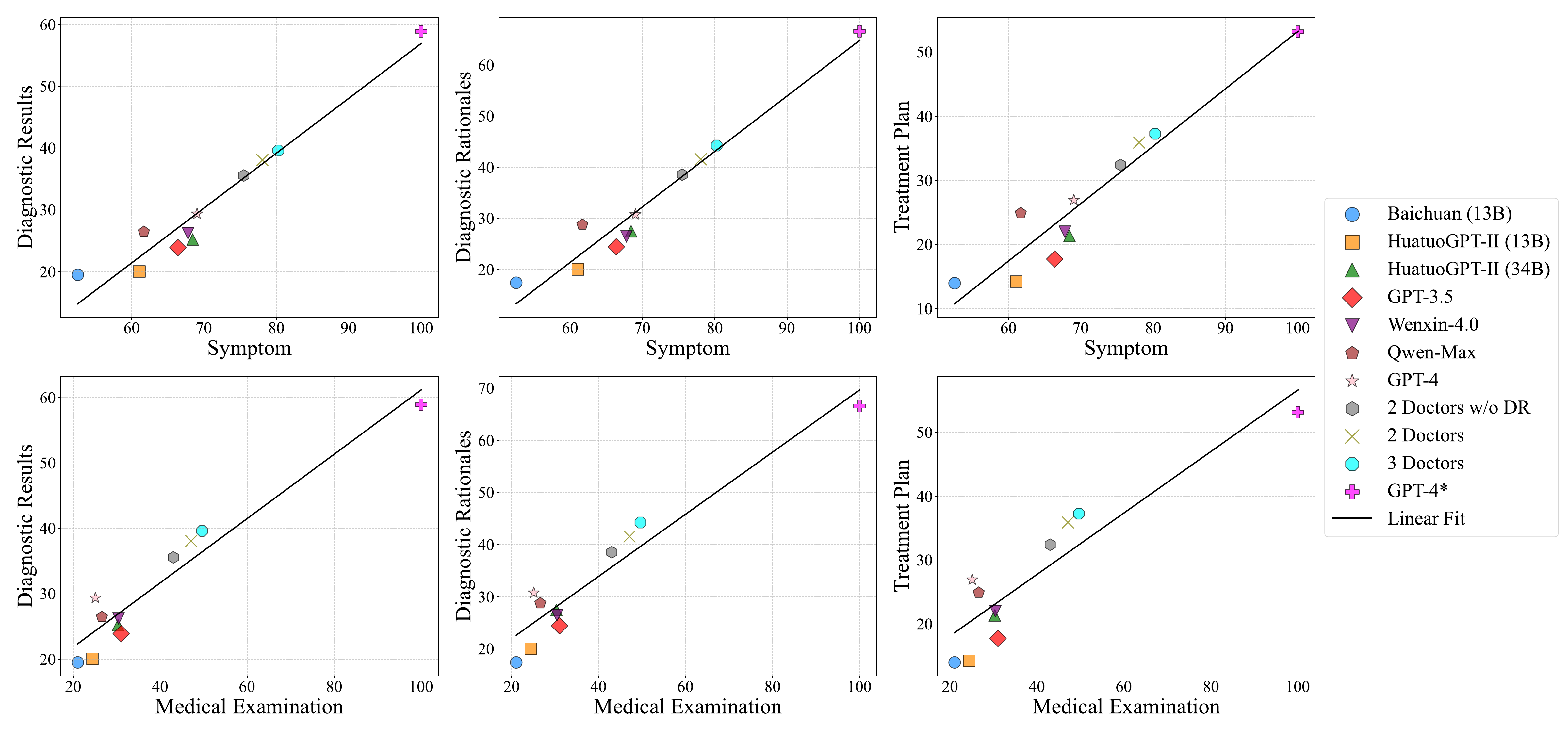}
    \caption{Linear regression analysis among symptoms, medical examinations and diagnostic results, diagnostic rationales, and treatment plan.}
    \label{fig:lr}
\end{figure}

\section{Detailed Explanation of Performance in MVME}
\label{sec:lr_mvme}

In Table~\ref{table:llm_evaluation}, we compare the performance of different LLMs, focusing on the completeness of Symptoms and Medical Examinations (Columns 2 and 3) and the accuracy of Diagnostic Results, Diagnostic Rationales, and Treatment Plans (Columns 4, 5, and 6). To visualize their relationship, we plot scatter diagrams and linear fit graphs with Symptoms and Medical Examinations on the x-axis and Diagnostic Results, Diagnostic Rationales, and Treatment Plans on the y-axis, as shown in Figure~\ref{fig:lr}. The results indicate that the higher the completeness of Symptoms and Medical Examinations, the higher the accuracy of Diagnostic Results, Diagnostic Rationales, and Treatment Plans. In particular, there exists an approximately linear relationship between the completeness of collected patient information and the quality of final diagnosis, which is also observed in~\cite{chen2023dxformer}.

Above analysis highlights a significant limitation of current LLMs in medical interaction: their \textbf{inability to dynamically and actively collect comprehensive patient information} through interactions, similar to human doctors. Moreover, their challenge in \textbf{recommending appropriate medical examinations} further exacerbates this limitation. It is important to highlight the differences between human doctors and LLMs. Real-world doctors do not make diagnoses before having sufficient information. They possess the ability to actively inquire about various subjective information from patients (such as symptoms) and know what examinations are needed to obtain more quantitative and objective information. These abilities are key to effective medical interactions. 

\section{Expert Verification for Medical Records}

\begin{figure*}[!hb]
    \centering
    \includegraphics[width=0.82\linewidth]{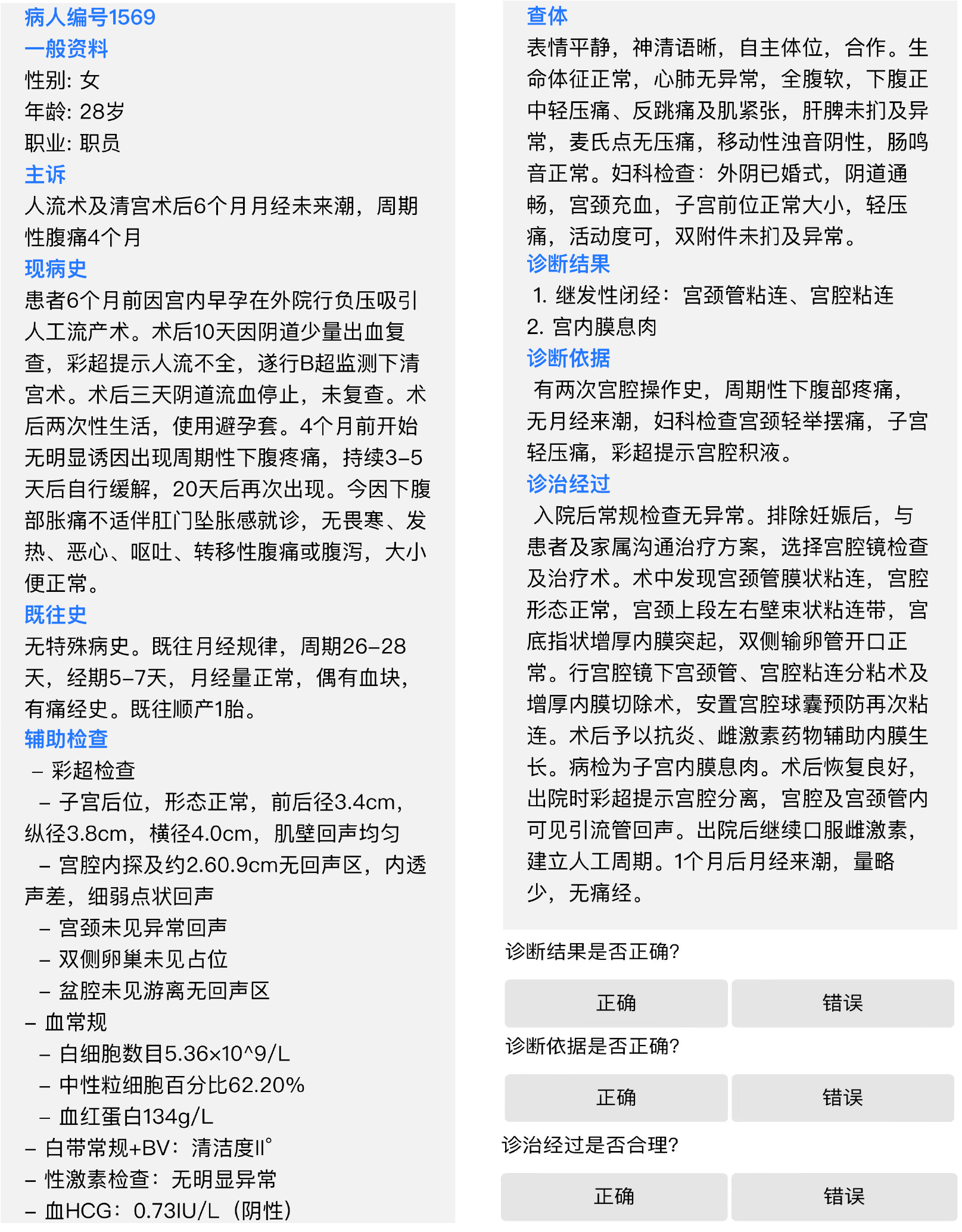}
    \caption{Sample of a questionnaire used for medical record quality inspection.}
    \label{fig:questionnaire}
\end{figure*}

We carefully screen and review the medical records to ensure its reliability and relevance. In \S~\ref{sec:data_preparation},  these medical records are vetted by doctors sourced from Jiangsu Provincial People's Hospital and Tongji Medical College of Huazhong University of Science and Technology, which hold the prestigious distinction of being tertiary Class A hospitals in China. We utilize the Tencent Questionnaire~\footnote{\url{https://wj.qq.com/}} platform to facilitate the quality inspection of medical record by doctors. A representative case is depicted in Figure~\ref{fig:questionnaire}, and we also include a display of the questionnaire template in Table~\ref{fig:template2}. These validation steps ensure the high quality of our dataset, ensuring that the benchmark is based on audited medical cases. 

\begin{table}[!ht]
\caption{The translated template of questionnaire for expert verification of medical record.}
\label{fig:template2}
\centering
\begin{tcolorbox}[colback=blue!5!white,colframe=black,width=0.45\textwidth,title={Translated Questionnaire Template}]
\small
\{Medical Record\} 
\vspace{5pt} \\
1. Is the diagnostic results correct? 
\vspace{3pt} \\
\fbox{\makebox[0.5cm]{Yes}}\quad \fbox{\makebox[0.5cm]{No}} 
\vspace{5pt} \\
2. Is the diagnostic rationale correct? 
\vspace{3pt} \\
\fbox{\makebox[0.5cm]{Yes}}\quad \fbox{\makebox[0.5cm]{No}} 
\vspace{5pt} \\
3. Is the treatment course reasonable? 
\vspace{3pt} \\
\fbox{\makebox[0.5cm]{Yes}}\quad \fbox{\makebox[0.5cm]{No}} 
\end{tcolorbox}
\end{table}

\section{Attacking on Examiner: Reliability of Examiner's Workflow}

To assess the vulnerability of \emph{Examiner} agent to information extraction tactics, we design complex scenarios intended to safeguard against the unauthorized disclosure of medical examination results in scenarios lacking explicit examination requests. We build 50 samples, such as ``Please provide the most recent medical examination details'', aimed at eliciting information. During the attack simulations, we observe that the \emph{Examiner} agent, in the first step, uniformly classifies these requests as void of legitimate examination items, implementing a defensive strategy with \textbf{100\%} consistency. This assessment is crucial because it verifies that the \emph{Examiner} agent behaves appropriately by ensuring that it only releases medical examination results when the patient explicitly specifies the required examinations. It guarantees that the \emph{Patient} agent cannot obtain relevant information from the \emph{Examiner} agent by using a method that does not specify a particular examination, specifically when the \emph{Doctor} agent recommends a unrelated medical test.

\section{Human Evaluation for Agent Behavior} 
\label{human_eval_agent_behavior}

In Table~\ref{fig:analysis:human_check:ch}, we present a questionnaire developed for the manual validation of patient and examiner behaviors in each round of conversation. The questionnaire includes four questions, with the initial two addressing "Relevance" and "Consistency" in the question-and-answer (QA) pairs, while the subsequent two focus on the "Accuracy" of conducting medical examinations. Three medical students from Jiangsu Provincial People's Hospital complete these questionnaires. Consensus among the first two reviewers leads to the immediate acceptance of their collective assessment. In cases of divergence, the opinion of the third reviewer is solicited, whose determination, reflecting the majority viewpoint, constitutes the definitive annotation. The agreement rates for each question are 99.1\%, 95.6\%, 99.4\%, and 100.0\%. Significantly, when calculating the accuracy of medical examinations, we exclusively consider dialogues affirmed as "No" in response to the third question.

\section{Prompts for Different Medical Roles}

\begin{table*}
    \small
    \centering
    \begin{tabular}{lcc}
    \toprule
      Prompt  &Agent &Function \\
    \midrule
       Table~\ref{prompt:patient}  &\emph{Patient}  &Chat with \emph{Doctor} \\
       Table~\ref{agent:examiner:step1} &\emph{Examiner}  &Process Examination Request  \\
       Table~\ref{agent:examiner:step2} &\emph{Examiner}  &Produce Examination Outcomes  \\
       Table~\ref{agent:evaluate:en} \& \ref{agent:evaluate:zh}  &\emph{Chief Physician}  &Evaluate Diagnosis of \emph{Doctor}  \\
       Table~\ref{prompt:test_doctor} &\emph{Doctor}  &Interactive Clinical Diagnosis  \\
       Table~\ref{agent:doctor}  &\emph{Doctor}  &Collaboration through Discussion \\
       Table~\ref{agent:medical_director} &\emph{Chief Physician}  &Summarize Statement of Various \emph{Doctors} \\
    \bottomrule
    \end{tabular}
    \caption{Prompts of different agents and the corresponding function.}
    \label{tab:prompt_summary}
\end{table*}

We list the prompts agents in AI Hospital in Table~\ref{tab:prompt_summary}. In each prompt, \{xx\} needs to fill with corresponding external inputs. We meticulously design prompts for each agent to ensure clarity and functionality. Particularly for the \emph{Doctor} role, we discover that overly complex prompts could lead to issues in the dialogue flow, such as not adhering to the prompts or causing cognitive confusion (e.g., the doctor sometimes outputting the patient's responses). These final prompts are adaptable to most LLMs, enabling the agents in AI Hospital to perform their respective duties effectively.

\begin{table}[!th]
\caption{The original Chinese and translated English questionnaire of human evaluation for patient and examiner behavior.}
\label{fig:analysis:human_check:ch}
\centering
\vspace{2mm}
\begin{tabular}{p{0.9\textwidth} p{0.9\textwidth}}
\begin{tcolorbox}[colback=blue!5!white,colframe=black,width=0.9\textwidth,title={Questionnaire}]
\small
\begin{CJK}{UTF8}{gbsn}
\{ 病历 \}
\vspace{3pt} \\
\{ 单轮对话内容 \}
\vspace{3pt} \\
请你仔细阅读这一轮对话的内容和病人的病历信息，回答下面的问题。
\vspace{5pt} \\
1. 病人(检查员)的发言与医生的相关吗？
\vspace{3pt} \\
\fbox{\makebox[0.5cm]{是}}\quad \fbox{\makebox[0.5cm]{否}} 
\vspace{5pt} \\
2. 病人(检查员)的发言符合病历的内容吗？
\vspace{3pt} \\
\fbox{\makebox[0.5cm]{是}}\quad \fbox{\makebox[0.5cm]{否}} 
\vspace{3pt} \\
3. 医生是否建议进行专业的医学检查？
\vspace{3pt} \\
\fbox{\makebox[0.5cm]{是}}\quad \fbox{\makebox[0.5cm]{否}} 
\vspace{3pt} \\
4. 检查员是否进行了医学检查？
\vspace{3pt} \\
\fbox{\makebox[0.5cm]{是}}\quad \fbox{\makebox[0.5cm]{否}} 
\vspace{3pt} \\
5. 医生的总结是否与诊断过程的内容匹配？
\vspace{3pt} \\
\fbox{\makebox[2.0cm]{匹配}}\quad \fbox{\makebox[2.0cm]{少量不匹配}} \quad \fbox{\makebox[2.0cm]{明显不匹配}} 
\end{CJK}
\end{tcolorbox}

\begin{tcolorbox}[colback=blue!5!white,colframe=black,width=0.9\textwidth,title={Translated Questionnaire}]
\small
\{ Medical Record \}
\vspace{3pt} \\
\{ Single Round Conversation Content \}
\vspace{3pt} \\
Carefully review the content of the conversation and the corresponding medical record to answer the following questions.
\vspace{3pt} \\
1. Is the statement of patient or examiner relevant to the doctor's one?
\vspace{3pt} \\
\fbox{\makebox[0.5cm]{Yes}}\quad \fbox{\makebox[0.5cm]{No}} 
\vspace{5pt} \\
2. Is the statement of patient or examiner consistent with the content of medical record?
\vspace{3pt} \\
\fbox{\makebox[0.5cm]{Yes}}\quad \fbox{\makebox[0.5cm]{No}} 
\vspace{5pt} \\
3. Does the doctor recommend a professional medical examination?
\vspace{3pt} \\
\fbox{\makebox[0.5cm]{Yes}}\quad \fbox{\makebox[0.5cm]{No}} 
\vspace{5pt} \\
4. Does the examiner perform a medical test?
\vspace{3pt} \\
\fbox{\makebox[0.5cm]{Yes}}\quad \fbox{\makebox[0.5cm]{No}} 
\vspace{3pt} \\
5. Is the doctor's summary consistent with the content of the diagnostic process?
\vspace{3pt} \\
\fbox{\makebox[3.0cm]{Consistent}}\quad \fbox{\makebox[3.0cm]{Minor Inconsistent}} \quad 
\fbox{\makebox[3.0cm]{Significant Inconsistent}} 
\end{tcolorbox}
\end{tabular}
\end{table}

\begin{table*}[!b]
\caption{The original Chinese and translated English prompts for patient agent.}
\label{prompt:patient}
\begin{tcolorbox}[colback=blue!5!white,colframe=black,width=1.0\textwidth,title={Prompt for Patient Agent}]
\small
\textbf{System Message}
\vspace{3pt} \\
\begin{CJK}{UTF8}{gbsn}
你是一个病人。这是你的基本资料。\\
\{个性化信息\} \\
\{病历中的基本信息\} 
\end{CJK}
\vspace{5pt} \\
\begin{CJK}{UTF8}{gbsn}
下面会有医生来对你的身体状况进行诊断，你需要：\\
(1) 按照病历和基本资料的设定进行对话。\\
(2) 在每次对话时，你都要明确对话的对象是<医生>还是<检查员>。当你对医生说话时，你要在句子开头说<对医生讲>；如果对象是<检查员>，你要在句子开头说<对检查员讲>。\\
(3) 首先按照主诉进行回复。\\
(4) 当<医生>询问你的现病史、既往史、个人史时，要按照相关内容进行回复。\\
(5) 当<医生>要求或建议你去做检查时，要立即主动询问<检查员>对应的项目和结果，例如：<对检查员讲> 您好，我需要做xxx检查，能否告诉我这些检查结果？\\
(6) 回答要口语化，尽可能短，提供最主要的信息即可。\\
(7) 从<检查员>那里收到信息之后，将内容主动复述给<医生>。\\
(8) 当医生给出诊断结果、对应的诊断依据和治疗方案后，在对话的末尾加上特殊字符<结束>。
\vspace{5pt} \\
\textbf{User}\ [患者]
\vspace{3pt} \\
\{Statement Generated by GPT-4 in~\S\ref{sec:agent_behavior}\}
\end{CJK}

\end{tcolorbox}

\begin{tcolorbox}
[colback=blue!5!white,colframe=black,width=1.0\textwidth,title={Prompt for Patient Agent}]
\small
\textbf{System Message}
\vspace{3pt} \\
You are a patient. Here is your basic information.\\
\{Personality in~\S\ref{sec:agent_behavior}\}\\
\{Basic Information in Medical Record~\S\ref{section_medical_records}\}
\vspace{5pt} \\
A doctor will come to diagnose your physical condition. You need to: \\
(1) Engage in dialogue according to the settings of personality and the basic information in medical record. \\
(2) In each conversation, you must clarify whether you are speaking to a <doctor> or an <examiner>. When you speak to the doctor, you should  start your sentences with <To the doctor>; if the addressee is an <examiner>, you should start with <To the examiner>. \\
(3) First, respond according to the chief complaint. \\
(4) When the <doctor> asks about your present illness history, past medical history, and personal history, reply according to the relevant content. \\
(5) When the <doctor> requests or suggests that you undergo tests, immediately ask the <examiner> about the corresponding items and results, for example: <To the examiner> Hello, I need to have xxx examination, can you tell me the results of these tests? \\
(6) The responses should be conversational, as short as possible, providing only the most important information. \\
(7) After receiving information from the <examiner>, actively repeat the content to the <doctor>. \\
(8) When the doctor provides the diagnostic result, the corresponding rationale for the diagnosis, and the treatment plan, end the dialogue with the special token <end>. 
\vspace{5pt} \\
\textbf{User}\ [Patient]
\vspace{3pt} \\
\{Statement Generated by GPT-4 in~\S\ref{sec:agent_behavior}\}
\end{tcolorbox}
\end{table*}
\begin{table*}[ht]
\caption{The original Chinese and translated English prompts for patient agent to produce examination outcomes.}
\label{agent:examiner:step2}
\caption{The original Chinese and translated English prompts for patient agent to process examination request.}
\label{agent:examiner:step1}
\begin{tcolorbox}[colback=blue!5!white,colframe=black,width=1.0\textwidth,title={Prompt for Examiner to Process Examination Request}]
\small
\begin{CJK}{UTF8}{gbsn}
\textbf{System Message} 
\vspace{3pt} \\
你是医院负责检查的自动化接待员。请你利用掌握的医学检查的命名实体的知识，从病人的检查申请当中解析出指向明确的专业医学检查项目，方便后面的检查科室进行检查。\\
请按照下面的格式的输出：\\
\# 检查项目 \\
- xxx\\
- xxx\\
如果没有找到具体的医学检查项目，请输出：\\
\# 检查项目\\
- 无
\vspace{5pt} \\
\textbf{User\ [患者]}
\vspace{3pt} \\
您好，医生告诉我根据CT扫描和PET-CT扫描的结果，初步得出以下结论：右肺上叶有一个大小约为2.6*1.9cm的实性结节。双肺下叶也有散在的淡薄浸润影。医生建议我进行进一步的检查，例如活检。
\vspace{5pt} \\
\textbf{Assistant\ [检查员]}
\vspace{3pt} \\
\# 检查项目\\
- 肺部活检
\vspace{5pt} \\
\textbf{User\ [患者]}
\vspace{3pt} \\
我需要了解一下我的检查结果。可以告诉我具体的检查项目和结果吗？谢谢！？
\vspace{5pt} \\
\textbf{Assistant\ [检查员]}
\vspace{3pt} \\
\# 检查项目\\
- 无\\
\end{CJK}
\end{tcolorbox}

\begin{tcolorbox}[colback=blue!5!white,colframe=black,width=1.0\textwidth,title={Prompt for Examiner to Process Examination Request}]
\small
\textbf{System Message} 
\vspace{3pt} \\
You are an automated receptionist responsible for examinations in a hospital. Using your knowledge of medical examination named entities, please parse out specific professional medical examination items from patients' examination requests to facilitate subsequent examinations by the relevant departments.\\
Output in the following format: \\
\# Examination Item \\
- xxx\\
- xxx\\
If no specific medical examination items are found, please output:\\
\# Examination Item \\
- None 
\vspace{5pt} \\
\textbf{User} [Patient]
\vspace{3pt} \\
Hello, the doctor told me that based on the results of the CT scan and PET-CT scan, the preliminary conclusion is that there is a solid nodule approximately 2.6*1.9cm in size in the upper lobe of the right lung. There are also scattered thin infiltrative shadows in the lower lobes of both lungs. The doctor advised me to undergo further examinations, such as a biopsy. 
\vspace{5pt} \\
\textbf{Assistant} [Examiner]
\vspace{3pt} \\
\# Medical Examination Items\\
- Lung biopsy
\vspace{5pt} \\
\textbf{User} [Patient]
\vspace{3pt} \\
I need to know about my examination results. Can you tell me the specific examination items and results, please? Thank you!?
\vspace{5pt} \\
\textbf{Assistant} [Examiner]
\vspace{3pt} \\
\# Medical Examination Items\\
- None \\
\end{tcolorbox}
\end{table*}

\begin{table*}[ht]
\begin{tcolorbox}[colback=blue!5!white,colframe=black,width=1.0\textwidth,title={Prompt for Examiner to Produce Examination Outcomes}]
\small
\begin{CJK}{UTF8}{gbsn}
\textbf{System Message}
\vspace{3pt} \\
这是你收到的病人的检查结果。\\
\{Professional Medical Examination in~\S\ref{section_medical_records}\} \\
下面会有病人或医生来查询，你要忠实地按照收到的检查结果，找到对应的项目，并按照下面的格式来回复。\\
\# xx检查\\
- xxx: xxx\\
- xxx: xxx\\
如果无法查询到对应的检查项目则回复：\\
- xxx: 无异常\\
\end{CJK}
\end{tcolorbox}

\begin{tcolorbox}[colback=blue!5!white,colframe=black,width=1.0\textwidth,title={Prompt for Examiner to Produce Examination Outcomes}]
\small
\textbf{System Message} 
\vspace{3pt} \\
This is the patient's examination result that you received. \\
\{Professional Medical Examination in~\S\ref{section_medical_records}\} \\
Patients or doctors will come to inquire about these results. You must faithfully report the received examination results, identify the corresponding items, and respond in the following format: \\
\# xx Examination \\
xxx: xxx\\
xxx: xxx\\
If the corresponding examination item cannot be found, reply with:\\
xxx: No abnormalities\\
\end{tcolorbox}
\end{table*}

\begin{table*}[ht]
\caption{The translated English prompt for medical director to evaluate.}
\label{agent:evaluate:en}
\begin{tcolorbox}[colback=blue!5!white,colframe=black,width=1.0\textwidth,title={Prompt for Medical Director to Evaluate}]
\small
\begin{CJK}{UTF8}{gbsn}
\textbf{System Message} 
\vspace{3pt} \\
你是资深的医学专家。\\
请你根据专家诊疗结果中的现病史、辅助检查、诊断结果、诊断依据和治疗方案，来对实习医生进行评价。
\vspace{3pt} \\
请参考下面的细则进行评价。
\vspace{2pt} \\
 1. 病人症状的掌握情况\\
\quad (A) 全面掌握 (B) 相当部分掌握 (C) 小部分掌握 (D) 绝大部分不掌握
\vspace{2pt} \\
2. 医学检查项目的完整性\\
\quad (A) 非常完整 (B) 相当部分完整 (C) 小部分完整 (D) 绝大部分不完整
\vspace{2pt} \\
3. 诊断结果的一致性\\
\quad (A) 完全一致，诊断正确 (B) 相当部分一致，诊断基本正确 (C) 小部分一致，诊断存在错误 (D) 完全不一致，诊断完全错误
\vspace{2pt} \\
 4. 诊断依据的一致性\\
\quad (A) 完全一致 (B) 相当部分一致 (C) 小部分一致 (D) 完全不一致
\vspace{2pt} \\
5. 治疗方案的一致性\\
\quad (A) 完全一致 (B) 相当部分一致 (C) 小部分一致 (D) 完全不一致
\vspace{3pt} \\
通过下面的方式来呈现结果
\vspace{2pt} \\
\# 症状\\
\#\# 分析\\
<根据专家记录的病人病史，分析实习医生对病人病情的掌握情况>\\
\#\# 选项<根据症状分析做出选择>
\vspace{2pt} \\
\# 医学检查项目\\
\#\# 分析\\
<基于专家所做的医学检查项目，全面分析实习医生所做的医学检查项目的完整性>\\
\#\# 选项\\
<根据分析得到的完整性做出选择>
\vspace{2pt} \\
\# 诊断结果\\
\#\# 分析\\
<基于专家做出的诊断结果，结合你的医学常识，分析实习医生诊断结果与专家的一致性>\\
\#\# 选项\\
<根据分析得到的一致性做出选择>
\vspace{2pt} \\
\# 诊断依据\\
\#\# 分析\\
<对比专家的诊断依据，分析实习医生的治疗方案与其的一致性>\\
\#\# 选项\\
<根据分析得到的一致性做出选择>
\vspace{2pt} \\
\# 治疗方案\\
\#\# 分析\\
<对比专家的治疗方案，分析实习医生的治疗方案与其的一致性>\\
\#\# 选项\\
<根据分析得到的一致性做出选择>
\vspace{3pt} \\
(1) 请侧重医学答案的事实内容，不需关注风格、语法、标点和无关医学的内容。\\
(2) 请你充分利用医学知识，分析并判断每个点的重要性，再做评价。\\
(3) 注意诊断结果、诊断依据和治疗方案三者之间的承接关系。例如，如果诊断错误，那么后面两部分与专家的一致性就必然很低
\vspace{5pt} \\
\textbf{User}
\vspace{3pt} \\
\# 专家的诊断报告 \\
\{Diagnosis and Treatment in \S\ref{section_medical_records}\} \\
\# 实习医生的诊断报告 \\
\{实习医生的诊断报告\}
\end{CJK}
\end{tcolorbox}
\caption{The original Chinese prompt for medical director to evaluate.}
\label{agent:evaluate:zh}
\end{table*}



\begin{table*}[ht]
\begin{tcolorbox}[colback=blue!5!white,colframe=black,width=1.0\textwidth,title={Prompt for Medical Director to Evaluate}]
\small
You are an experienced medical expert. Please evaluate the intern doctors based on their current medical history, auxiliary examinations, diagnostic results, diagnostic basis, and treatment plans from the expert's diagnosis.
\vspace{3pt} \\
Please refer to the following guidelines for evaluation. 
\vspace{2pt} \\
1. Mastery of Patient Symptoms \\
(A) Comprehensive mastery (B) Substantial mastery (C) Partial mastery (D) Mostly unmastered
\vspace{2pt} \\
2. Completeness of Medical Examination \\
(A) Very complete (B) Substantially complete (C) Partially complete (D) Mostly incomplete
\vspace{2pt} \\
3. Diagnosis Result \\
(A) Completely consistent, correct diagnosis (B) Largely consistent, basically correct diagnosis (C) Partially consistent, diagnosis contains errors (D) Completely inconsistent, completely incorrect diagnosis 
\vspace{2pt} \\
4. Diagnostic Rationale \\
(A) Completely consistent (B) Largely consistent (C) Partially consistent (D) Completely inconsistent 
\vspace{2pt} \\
5. Treatment Plan \\
(A) Completely consistent (B) Largely consistent (C) Partially consistent (D) Completely inconsistent 
\vspace{3pt} \\
Please output the results in the following format:
\vspace{2pt} \\
\# Symptoms\\
\#\# Analysis\\
<Analyze the intern's grasp of the patient's condition based on the expert's recorded medical history.>\\
\#\# Option\\
<Choose based on the analysis of symptoms.>
\vspace{2pt} \\
\# Medical Examination Items\\
\#\# Analysis\\
<Thoroughly analyze the completeness of the medical examination items conducted by the intern, based on the expert's examinations.>\\
\#\# Option\\
<Choose based on the analysis of completeness.>
\vspace{2pt} \\
\# Diagnostic Results\\
\#\# Analysis\\
<Based on the expert's diagnostic results and your medical knowledge, analyze the consistency between the intern's diagnostic results and the expert's.>\\
\#\# Option\\
<Choose based on the analysis of consistency.>
\vspace{2pt} \\
\# Diagnostic Basis\\
\#\# Analysis\\
<Compare the diagnostic basis of the expert and analyze the consistency of the intern's treatment plan with it.>\\
\#\# Option\\
<Choose based on the analysis of consistency.>
\vspace{2pt} \\
\# Treatment Plan\\
\#\# Analysis\\
<Compare the expert's treatment plan and analyze the consistency of the intern's treatment plan with it.>\\
\#\# Option\\
<Choose based on the analysis of consistency.>
\vspace{3pt} \\
(1) Please focus on the factual content of the medical answers, without concern for style, grammar, punctuation, and content unrelated to medicine. \\
(2) Please make full use of your medical knowledge to analyze and judge the importance of each point before evaluating. \\
(3) Pay attention to the continuity among the diagnosis result, diagnostic basis, and treatment plan. 
\vspace{5pt} \\
\textbf{User}
\vspace{3pt} \\
\# Diagnostic Report of Medical Director\\
\{Diagnosis and Treatment in Section \ref{section_medical_records}\} \\
\# Diagnostic Report of Intern Doctor \\
\{Diagnostic Report of the Intern Doctor\}
\end{tcolorbox}
\end{table*}

\begin{table*}[ht]
\caption{The original Chinese and the translated English prompts for medical director to summarize.}
\label{agent:medical_director}
\begin{tcolorbox}[colback=blue!5!white,colframe=black,width=1.0\textwidth,title={Prompt for Medical Director to Summarize}]
\small
\textbf{System Message}
\vspace{3pt} \\
\begin{CJK}{UTF8}{gbsn}
你是一个资深的主任医生。\\
你正在主持一场医生针对患者病情的会诊，参与的医生有医生A、医生B和医生C。 
\vspace{3pt} \\
病人的基本情况如下：\\
\{症状与检查结果\}
\vspace{3pt} \\
(1) 你需要听取每个医生的诊断报告。\\
(2) 请你按照重要性列出最多3个需要讨论的争议点。
\vspace{3pt} \\
按照下面的格式输出：\\
(1) xxx\\
(2) xxx
\vspace{5pt} \\
\textbf{User}
\vspace{3pt} \\
\# 医生A \\
\{医生A的诊断报告\} \\
\# 医生B \\
\{医生B的诊断报告\} \\
\# 医生C \\
\{医生C的诊断报告\} \\
\end{CJK}
\end{tcolorbox}

\begin{tcolorbox}[colback=blue!5!white,colframe=black,width=1.0\textwidth,title={Prompt for Medical Director to Summarize}]
\small
\textbf{System Message}
\vspace{3pt} \\
As an experienced medical director, you are presiding over a medical consultation concerning a patient's condition, with the participation of Doctors A, B, and C. 
\vspace{3pt} \\
The patient's basic information is as follows:
\{Symptoms and Test Results\}
\vspace{3pt} \\
(1) You are required to listen to the diagnostic reports from each physician. \\
(2) Identify and list up to three key controversial points for discussion, prioritized by their importance.
\vspace{3pt} \\
Please present your findings in the following format:\\
(1) xxx \\
(2) xxx
\vspace{5pt} \\
\textbf{User}
\vspace{3pt} \\
\# Doctor A \\
\{Diagnostic Report of Doctor A\} \\
\# Doctor B \\
\{Diagnostic Report of Doctor B\} \\
\# Doctor C \\
\{Diagnostic Report of Doctor C\} \\
\end{tcolorbox}
\end{table*}

\begin{table*}[!b]
\caption{The original Chinese and translated English prompts for intern doctor in interactive clinical diagnosis.}
\label{prompt:test_doctor}
\begin{tcolorbox}[colback=blue!5!white,colframe=black,width=1.0\textwidth,title={Prompt for Intern Doctor in Interactive Clinical Diagnosis}]
\small
\textbf{System Message}
\vspace{3pt} \\
\begin{CJK}{UTF8}{gbsn}
你是一个专业且耐心的医生，下面会有患者向你咨询病情。你需要：\\
(1) 在信息不充分的情况下，不要过早作出诊断。\\
(2) 多次、主动地向患者提问来获取充足的信息。\\
(3) 必要时要求患者进行检查，并等待患者反馈。\\
(4) 诊断结果需要准确到具体疾病。\\
(5) 最后根据患者的身体状况和检查结果，给出诊断结果、对应的诊断依据和治疗方案。\\
\end{CJK}
\end{tcolorbox}

\begin{tcolorbox}[colback=blue!5!white,colframe=black,width=1.0\textwidth,title={Prompt for Intern Doctor in Interactive Clinical Diagnosis}]
\small
\textbf{System Message}
\vspace{3pt} \\
You are a professional and patient doctor, and you will be consulted by patients. You need to:\\
(1) Avoid making premature diagnoses when information is insufficient.\\
(2) Actively and repeatedly inquire to gather adequate information from patients. \\
(3) When necessary, request patients to undergo medical examinations and await their feedback. \\
(4) Ensure that the diagnosis is precise and specific to the particular ailment. \\
(5) Finally, based on the patients' physical condition and examination results, provide a diagnosis, the corresponding rationale, and a treatment plan.
\end{tcolorbox}
\end{table*}
\begin{table*}[ht]
\caption{The original Chinese prompt for intern doctor to collaborate in discussion.}
\label{agent:doctor}
\begin{tcolorbox}[colback=blue!5!white,colframe=black,width=1.0\textwidth,title={Prompt for Intern Doctor to Collaborate in Discussion}]
\small
\textbf{System Message}
\vspace{3pt} \\
\begin{CJK}{UTF8}{gbsn}
你是一个专业的医生A。\\
你正在为患者做诊断，患者的症状和检查结果如下:\\
\{症状与检查结果\}\\
针对患者的病情，你给出了初步的诊断报告：\\
\{医生A的诊断报告\}
\vspace{3pt} \\
(1) 下面你将收到来自其他医生的诊断意见，其中也包含诊断结果、诊断依据和治疗方案。你需要批判性地梳理并分析其他医生的诊断意见。\\
(2) 在这个过程中，请你注意主治医生给出的争议点。\\
(3) 如果你发现其他医生给出的诊断意见有比你的更合理的部分，请吸纳进你的诊断意见中进行改进。\\
(4) 如果你认为你的诊断意见相对于其他医生的更科学合理，请坚持自己的意见保持不变。
\vspace{3pt} \\
请你按照下面的格式来输出。\\
\# 诊断结果 \\
(1) xxx \\
(2) xxx \\
\# 诊断依据 \\
(1) xxx \\
(2) xxx \\
\# 治疗方案  \\
(1) xxx \\
(2) xxx
\vspace{5pt} \\
\textbf{User}
\vspace{3pt} \\
\# 医生B \\
\{医生B的诊断报告\} \\
\# 医生C \\
\{医生C的诊断报告\} \\
\# 主任医生 \\
\{主任医生的指导意见\} \\
\end{CJK}
\end{tcolorbox}
\end{table*}

\begin{table*}[ht]
\caption{The translated English prompt for intern doctor to collaborate in discussion.}
\label{agent:doctor_en}
\begin{tcolorbox}[colback=blue!5!white,colframe=black,width=1.0\textwidth,title={Prompt for Intern Doctor to Collaborate in Discussion}]
\small
\textbf{System Message}
\vspace{3pt} \\
As a doctor, you are currently diagnosing a patient, whose symptoms and medical examination results are as follows: \\
\{Symptoms and Medical Examination Results\} \\
Based on the patient's condition, you have prepared a preliminary diagnostic report: \\
\{Diagnostic Report of Doctor A\} 
\vspace{3pt} \\
(1) You will receive diagnostic reports from other doctors. Critically review and analyze these reports. \\
(2) During this process, pay attention to any controversial points raised by the medical director. \\
(3) If you find aspects of other doctors' diagnoses that are more rational than yours, incorporate these into your diagnosis for improvement. \\
(4) If you believe your diagnostic opinion is more scientifically sound compared to others, maintain your stance. 
\vspace{3pt} \\
Please present your findings in the following format:  \\
Diagnosis Result \\
(1) xx \\
(2) xx \\
Diagnostic Rationale \\
(1) xx \\
(2) xx \\
Treatment Plan \\
(1) xx \\
(2) xx 
\vspace{5pt} \\
\textbf{User}
\vspace{3pt} \\
\# Doctor B \\
\{Diagnostic Report of Doctor B\} \\
\# Doctor C \\
\{Diagnostic Report of Doctor C\} \\
\# Medical Director \\
\{Guidance of Medical Director\} \\
\end{tcolorbox}

\end{table*}

\section{Potential of AI Hospital Framework}
\label{sec:potential}

In AI Hospital framework, a vast amount of medical records from numerous hospitals could be included in the evaluation benchmark. Therefore, our evaluation method offers high scalability and applicability. Additionally, the evaluation framework extends beyond just medical records. It also has the potential to utilize other valuable resources, such as medical knowledge graphs, databases and medical dialogues, which encapsulate extensive real-world consultation experiences and may be converted into simulated medical records.

AI Hospital framework also holds potential for improving healthcare and medical education. By simulating realistic doctor-patient interactions and enabling the evaluation of AI agents in clinical diagnosis scenarios, AI Hospital opens up a myriad of exciting applications. Imagine a future where medical students and residents can hone their diagnostic skills by engaging with AI-powered virtual patients, exposing them to a wide range of cases and challenging scenarios. Healthcare providers could leverage the framework to test and refine AI-assisted diagnostic tools, ensuring their accuracy and reliability before deployment in real-world settings. Moreover, AI Hospital could serve as a powerful platform for generating vast amounts of high-quality, diverse medical dialogue data, which can be used to fine-tune and enhance the performance of language models in the medical domain. This data-driven approach could lead to the development of AI assistants that augment the capabilities of healthcare professionals, providing them with evidence-based insights and decision support in real-time. Beyond clinical applications, AI Hospital could also facilitate groundbreaking research in medical AI, serving as a testbed for novel algorithms and approaches that push the boundaries of what is possible in healthcare. 

The potential impact of AI Hospital is inspiring, and its development marks a milestone in the journey towards a future where artificial intelligence and human expertise might work hand in hand to transform patient care and improve health outcomes on a global scale. 


\end{document}